\documentclass[letterpaper]{article} 
\usepackage[preprint]{aaai2027}
\usepackage[hyphens]{url}  
\usepackage{graphicx} 
\usepackage{natbib}  
\usepackage{caption} 
\usepackage{algorithm}
\usepackage{algorithmic}
\usepackage{amsmath}
\usepackage{amssymb}
\usepackage{newfloat}
\usepackage{listings}
\DeclareCaptionStyle{ruled}{labelfont=normalfont,labelsep=colon,strut=off} 
\floatstyle{ruled}
\newfloat{listing}{tb}{lst}{}
\floatname{listing}{Listing}

\usepackage{booktabs}
\usepackage[table]{xcolor}
\usepackage{graphicx}
\usepackage{xspace}
\usepackage{arydshln}
\newcommand{\ours}{\textsc{MACRO}\xspace}

\newcommand{\std}[1]{\textsubscript{\textcolor{gray}{\(\pm\)#1}}}

\definecolor{HeaderSoft}{HTML}{E8EEF7}
\definecolor{HeaderText}{HTML}{1F2937}
\definecolor{SoftRule}{HTML}{9CA3AF}

\definecolor{QwenBlue}{HTML}{4F6BED}
\definecolor{QwenTeal}{HTML}{0F9E9E}
\definecolor{LlamaPurple}{HTML}{8B5CF6}
\definecolor{OlmoGreen}{HTML}{2E9F75}
\definecolor{QwenOrange}{HTML}{F97316}
\definecolor{QwenRed}{HTML}{DC2626}
\definecolor{MistralRose}{HTML}{E11D48}
\definecolor{DeepSeekCyan}{HTML}{0284C7}

\definecolor{ModelBarGray}{HTML}{4B5563}
\newcommand{\modelbar}[2]{%
\rowcolor{ModelBarGray}
\multicolumn{15}{l}{\textcolor{white}{\textbf{Model: #2}}} \\
}

\definecolor{modelgray}{HTML}{EFEFEF}
\definecolor{lightyellow}{HTML}{FFF7A8}
\definecolor{lightorange}{HTML}{F6C28B}
\definecolor{lightred}{HTML}{F58A8A}

\def\redc{\cellcolor[HTML]{FF999A}}
\def\orangec{\cellcolor[HTML]{FFCC99}}
\def\yellowc{\cellcolor[HTML]{FFF8AD}}

\newcommand{\kept}{\textcolor{OlmoGreen}{\(\checkmark\)}}
\newcommand{\dropped}{\textcolor{lightred}{\(\times\)}}
\newcommand{\Ours}{MACRO}
\title{MACRO: \underline{Ma}rkov \underline{C}hain \underline{Ro}uting of Transformer Layers}
\author{
    Pawe\l{} Batorski\textsuperscript{\rm 1},
    Abtin Pourhadi\textsuperscript{\rm 1},
    Akylgali Aitaza\textsuperscript{\rm 1},
    Przemys\l{}aw Spurek\textsuperscript{\rm 2,\rm 3},
    Paul Swoboda\textsuperscript{\rm 1}
}
\affiliations{
    \textsuperscript{\rm 1}Heinrich Heine University D\"usseldorf\\
    \textsuperscript{\rm 2}Jagiellonian University\\
    \textsuperscript{\rm 3}IDEAS Research Institute
}

\makeatletter
\let\mrtl@aaai@maketitle\@maketitle
\renewcommand{\@maketitle}{%
  \mrtl@aaai@maketitle
  \vspace{-0.8em}
  \begin{center}
    \begin{minipage}{0.99\textwidth}
      \includegraphics[height=0.36\linewidth,keepaspectratio]{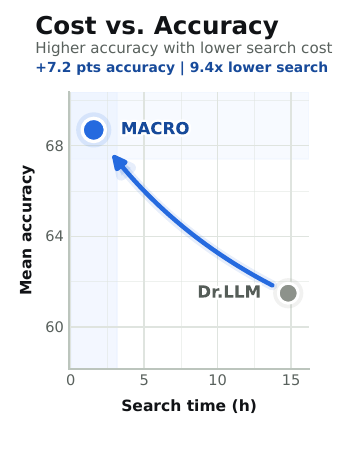}\hfill
      \includegraphics[height=0.36\linewidth,keepaspectratio]{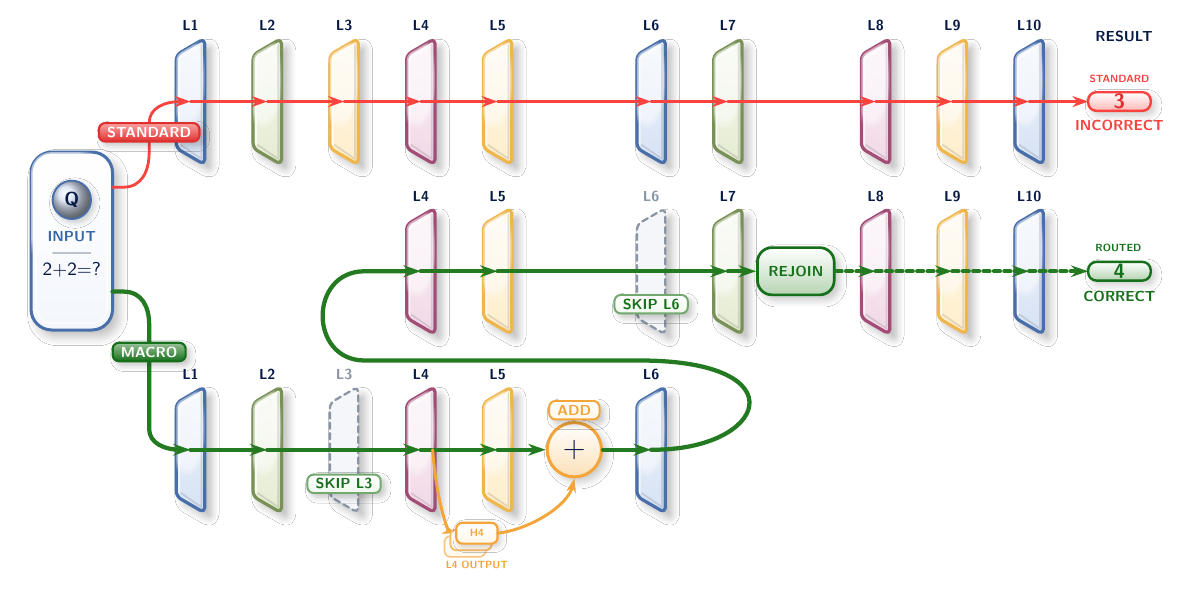}
      \captionof{figure}{Against Dr.LLM, MACRO improves mean accuracy while reducing route-search time from 14.8 to 1.6 hours per benchmark, averaged over six shared model settings.}
    \end{minipage}
  \end{center}
  \vspace{-0.5em}
}
\makeatother

\begin{document}

\maketitle

\begin{abstract}
Standard Large Language Models (LLMs) execute layers sequentially. Dynamic layer routing, i.e.\ search for a different execution path through layers involving layer repetitions, skips and other moves, can improve performance.
Existing routing approaches often require updating model weights, running expensive search loops per test instance, or demand ground-truth labels during inference. 
In this work, we propose \textbf{Markov Chain Routing of Transformer Layers (\Ours{})}, a framework that learns task-specific routes over LLM architectures without modifying underlying parameters. 
\Ours{} models layer routing as a context-dependent Markov policy conditioned on layer indices, computation budget phases, directional displacements, and operator context, supporting skip, repeat, and residual hidden-state addition operations. 
The Markov route distribution is updated via feedback on training data and decoded using a top-$k$ Viterbi algorithm to isolate high-probability candidate programs. 
We evaluate \Ours{} across diverse reasoning and knowledge benchmarks on multiple open-weight LLMs.
\Ours{} achieves a \textbf{+5.0\%} average accuracy improvement over the unrouted baselines, with largest gains on small models.
We outperform the best dynamic routing approach Dr.\ LLM  by \textbf{+7.2\%}, while reducing route-search time \textbf{9.4$\times$} (from 14.8 to 1.6 hours).
Our code is publicly available at \url{https://github.com/Batorskq/MACRO}.
\end{abstract}



\section{Introduction}

\begin{figure}[t]
\centering
\includegraphics[width=\columnwidth]{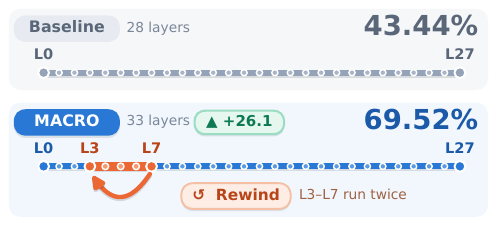}
\caption{A single \ours{} route on Qwen3-1.7B for GSM8K. The baseline executes
the 28 transformer blocks once, front to back. The searched route runs
\texttt{L0}--\texttt{L7}, rewinds to \texttt{L3}, and then continues to
\texttt{L27}, so \texttt{L3}--\texttt{L7} execute twice and 33 block calls are
made in total. The weights are frozen and only the execution order changes.}
\label{fig:qwen17b_paths}
\end{figure}

Standard Large Language Models (LLMs) operate on a rigid computational schedule, executing transformer blocks sequentially from the first layer to the last. 
This approach ignores the potential for dynamic computation pathways.
Prior work has explored adaptive depth primarily to accelerate inference via early exits \citep{teerapittayanon2016branchynet,xin-etal-2020-deebert,zhou2020bert} or structured layer dropping \citep{Fan2020Reducing,elhoushi-etal-2024-layerskip}. 
However, actively restructuring the execution route of a frozen model to \textit{improve reasoning accuracy} on complex tasks remains a highly challenging and relatively unexplored frontier.
Figure~\ref{fig:qwen17b_paths} illustrates what such a restructuring can buy: on GSM8K, rewinding Qwen3-1.7B from layer~7 back to layer~3 and replaying five blocks raises test accuracy from 43.4\% to 69.5\%, without changing a single weight.

Recent efforts to achieve accuracy gains through layer routing highlight significant practical bottlenecks. Dr.LLM \citep{heakl2026drllm} bypasses the need for test-time labels by training lightweight routers, yet it necessitates an expensive offline Monte Carlo Tree Search (MCTS)-driven data generation and training pipeline. Furthermore, purely exhaustive approaches, such as Neuroanatomy \cite{ng2026rys}, operate in a constrained search space and are computationally prohibitive for per-task adaptation or search over more complex routes. 

To overcome these limitations, we introduce \textbf{Markov Chain Routing of Transformer Layers (MACRO)}. 
We formalize layer routing as a discrete, context-dependent stochastic process.
We choose Markov chains \citep{norris1998markov} for the following reasons:
First, unlike deep reinforcement learning or auxiliary neural routers, a Markov transition matrix is lightweight and requires no parameter updates or backpropagation through the LLM.
Second, it naturally mitigates overfitting on small validation datasets by structurally constraining the decision space. Most importantly, modeling the routing as a Markov chain allows us to leverage exact, polynomial-time sequence decoding (the Viterbi algorithm) instead of relying on stochastic sampling or computationally expensive tree searches at test time.

\Ours{} operates through on a compact state and action space: The state space encompasses the current layer, the remaining computation budget, the incoming displacement and the operator context.
From any state, the policy can execute feasible actions, which include moving to a nearby layer, adding a previous hidden state (drawing inspiration from recurrent-depth behaviors \citep{chen2025inner,bae2026mixture}), or rejoining the standard suffix.
Rather than employing per-sample routing, the framework learns a single, task-specific Markov transition distribution.
This global policy is iteratively updated using an estimation-of-distribution approach on a subset of training data, augmented by a replay buffer to ensure stable convergence.
Once the Markov chain is fully trained, exploration is completely halted.
At this final stage, we employ a top-$k$ Viterbi algorithm to deterministically extract the highest-probability valid routing programs, selecting the absolute best-performing route via a held-out validation set for uniform test-time deployment.

In summary, the main contributions of this work are:
\begin{itemize}
    \item We propose \Ours{}, a parameter-efficient framework for adaptive layer routing in frozen and quantized LLMs, modeled as a context-dependent Markov policy.
    \item We introduce an exact top-$k$ Viterbi decoding scheme equipped with structural feasibility masks, which eliminates the need for per-sample test-time search, auxiliary neural network training, or ground-truth inference oracles.
    \item We demonstrate empirically that \Ours{} consistently outperforms prior dynamic routing methods across diverse reasoning and knowledge benchmarks, yielding a \textbf{+5.0\%} average accuracy improvement over the non-routed baseline, \textbf{+7.2\%} over Dr.\ LLM, and \textbf{+1.5\%} over Dr.\ LLM with an equivalent action space, while reducing the route-search computational cost by \textbf{9.4$\times$}.
\end{itemize}

\section{Related Work}

\paragraph{Adaptive Depth and Layer Routing}
Much prior work exploits redundancy in transformer depth mainly to make
inference faster through early exits
\citep{teerapittayanon2016branchynet,xin-etal-2020-deebert,liu-etal-2020-fastbert,zhou2020bert,elbayad2020depth,liu2021faster,hou2020dynabert,schuster2022calm,chen2024eellm,jazbec2024fast},
structured layer dropping
\citep{Fan2020Reducing,liu-etal-2021-ebert,men2025shortgpt,zhao2025skipgpt}, and
LayerSkip \citep{elhoushi-etal-2024-layerskip}. Mixture-of-Depths
\citep{raposo2024mixture}, adaptive layer skipping \citep{luo2025adaptive}, and
router tuning \citep{he-etal-2025-router} similarly adapt which layers or tokens
receive computation, while earlier dynamic networks learn input-dependent
execution paths \citep{wu2018blockdrop,wang2018skipnet}. Our objective differs:
we use layer routing to improve accuracy while keeping the pretrained model
fixed.

The closest works are Dr.LLM \citep{heakl2026drllm}, CoLA \citep{li2025cola},
and Neuroanatomy~\citep{ng2026rys}. Dr.LLM trains lightweight routers from
MCTS-derived supervision, requiring route-label generation and router training.
CoLA performs per-sample MCTS over skipped and looped layers with ground-truth
reward, and Neuroanatomy exhaustively searches a limited route space. \ours{}
learns one Markov route distribution per task from train/validation feedback and
decodes it with top-\(k\) Viterbi, avoiding per-test ground-truth search and any
model-weight updates.

\paragraph{Recurrent and Expert Routing}
Universal Transformers \citep{dehghani2018universal}, looped transformers
\citep{giannou2023looped,yang2024looped,zhu2025scaling}, and recurrent-depth
models \citep{chen2025inner,geiping2026scaling,bae2026mixture} show that
repeated computation over depth can support algorithmic or reasoning behavior.
Mixture-of-experts models \citep{shazeer2017,lepikhin2021gshard,fedus2022switch}
and routing experts \citep{wu2025routing} route tokens across parameter experts;
in contrast, our experts are the existing blocks of a single pretrained
transformer. The lightweight Markov policy \citep{norris1998markov},
feasibility masks, and top-\(k\) Viterbi decoding remain compatible with frozen
and quantized LLMs.

\begin{table}[!tb]
\centering
\footnotesize
\setlength{\tabcolsep}{1.7pt}
\renewcommand{\arraystretch}{1.18}
\arrayrulecolor{SoftRule}
\begin{tabular}{@{}lccccc@{}}
\toprule
\rowcolor{HeaderSoft}
\textcolor{HeaderText}{\textbf{Method}}
& \textcolor{HeaderText}{\textbf{Acc.}}
& \textcolor{HeaderText}{\textbf{No Labels}}
& \textcolor{HeaderText}{\textbf{No Per-Ex.}}
& \textcolor{HeaderText}{\textbf{No Pre-Gen}}
& \textcolor{HeaderText}{\textbf{Frozen}} \\
\midrule
Early exit/Skip & \dropped & \kept & \kept & \kept & \dropped \\
CoLA            & \kept & \dropped & \dropped & \kept & \kept \\
Dr.LLM          & \kept & \kept & \kept & \dropped & \kept \\
\rowcolor{QwenBlue!8}
\textbf{\ours{}}& \kept & \kept & \kept & \kept & \kept \\
\bottomrule
\end{tabular}
\caption{Qualitative comparison of layer-adaptation methods. ``Acc.'' denotes
an explicit accuracy-improvement objective; ``No Labels'' means no test labels
are used; ``No Per-Ex.'' means no per-example test-time search is required;
``No Pre-Gen'' means no offline route-label generation stage is required.}
\label{tab:method_comparison}
\end{table}

\section{Method}

\begin{figure*}[!t]
  \centering
  \includegraphics[width=\textwidth]{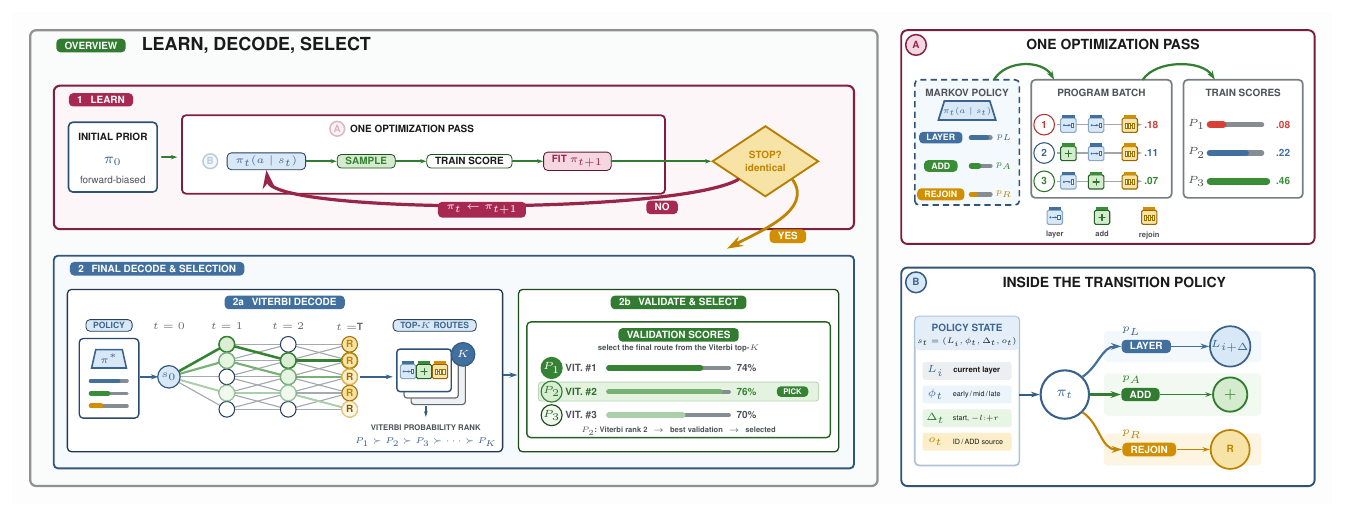}
  \caption{
  The left panel summarizes the complete workflow. \textbf{(1)} Starting from a forward-biased prior~\(\pi_0\), each iteration samples \(N\) routing programs, scores them by executing the frozen model on training data, and refits the policy using the top-performing programs, weighted exponentially by training score, until convergence or the iteration limit. \textbf{(2a)} Top-\(K\) Viterbi decoding ranks feasible programs under the learned policy as \(P_1 \succ P_2 \succ \cdots \succ P_K\). \textbf{(2b)} Held-out validation selects among the execution-distinct candidates independently of their Viterbi rank. Panel \textbf{(A)} expands the sampling, scoring, and refitting steps. Panel \textbf{(B)} shows the masked categorical policy conditioned on layer, budget phase, incoming displacement, and operation provenance, over local LAYER, parameterized ADD, and deterministic REJOIN actions.
  }
  \label{fig:method}
\end{figure*}

\paragraph{Problem Statement}

Consider a pretrained transformer with frozen weights, whose forward pass
applies $L$ layer blocks $f_1, \dots, f_L$ in a fixed order. A route $r$
relaxes this schedule into a program over the blocks. At each step, it chooses
which block to apply, so blocks may be skipped, repeated, or revisited by going
back to earlier ones. The route may also add an earlier hidden state to the
running hidden state before applying the next block. Let $\mathcal{R}$ be the
space of such routes and let $M_r$ denote the model run under route $r$. Given
a dataset
$\mathcal{D} = \{(x_i, y_i)\}_{i=1}^{N}$, we seek the route that maximizes
accuracy:
\begin{equation}
\begin{aligned}
  r^{\star}
  &= \operatorname*{arg\,max}_{r \in \mathcal{R}} \operatorname{Acc}(r), \\
  \operatorname{Acc}(r)
  &= \frac{1}{N} \sum_{i=1}^{N}
  \mathbf{1}\{ M_r(x_i) = y_i \}.
\end{aligned}
\end{equation}

\paragraph{Dataset Split}
Given a benchmark dataset $\mathcal{D}$, we partition it into three disjoint subsets: a training set $\mathcal{D}_{\mathrm{train}}$, a validation set $\mathcal{D}_{\mathrm{val}}$, and a test set $\mathcal{D}_{\mathrm{test}}$:
\[
\mathcal{D}
=
\mathcal{D}_{\mathrm{train}}
\cup
\mathcal{D}_{\mathrm{val}}
\cup
\mathcal{D}_{\mathrm{test}}.
\]
The training and validation sets are used during routing search, whereas the test set is held out and used only for final evaluation. Throughout this procedure, the model parameters remain fixed, and optimization is performed over the space of routings. To reduce overfitting during search, we further partition the training set into $I$ disjoint subsets,
\[
\mathcal{D}_{\mathrm{train}}
=
\bigcup_{i=1}^{I} \mathcal{D}_i,
\]
which are used across search rounds to evaluate candidate routes on fresh subsets of training data. At search iteration $i$, candidates are evaluated on the current subset together with a small replay buffer of examples from previous rounds. We denote the buffer by $\mathcal{B}$, initialize it as $\mathcal{B}_0=\emptyset$, and use
\[
\mathcal{S}_i = \mathcal{D}_i \cup \mathcal{B}_{i-1}
\]
as the route-evaluation set. After each iteration, the buffer is updated by adding a randomly sampled fraction \(\lambda_{\mathrm{rep}}\) of the current subset:
\[
\mathcal{B}_i
=
\mathcal{B}_{i-1}
\cup
\mathrm{Sample}_{\lambda_{\mathrm{rep}}}(\mathcal{D}_i).
\]
This schedule keeps each round focused mainly on fresh data while preserving limited coverage of examples seen in earlier iterations.

\paragraph{Markov Route Generator}
We model route construction as a context-dependent Markov policy
\(\pi_\theta(a_t \mid s_t)\). A route is represented as a typed program that
starts at the first transformer layer and is extended one action at a time until
a terminal \textsc{Rejoin} action is selected. The state \(s_t\) is a compact
summary of the partial program, including the current layer, the current
length or budget phase, and optional context such as the previous layer, the
incoming layer displacement, and whether the current hidden state was produced
by a plain layer or by an add operation. At each step, infeasible actions are
masked so that the program stays within layer bounds and the maximum route
length.

\paragraph{State Space}
For a partial program, the state factorizes as $s_t=(\ell,\phi,\delta,o)$, and
the next-action distribution is conditioned on all four components.
\begin{enumerate}
    \item The \emph{current layer} $\ell\in\{0,\dots,L-1\}$ is the most recently produced layer in the partial program.
    \item The \emph{budget phase} $\phi\in\{\textsc{early},\textsc{mid},\textsc{late}\}$ buckets the fraction of remaining route-length budget into three bins, biasing the policy toward \textsc{Rejoin} as the route nears its maximum length. 
    \item The \emph{incoming displacement} $\delta=\ell-\ell_{\mathrm{prev}}\in\{-r,\dots,r\}$, together with a dedicated start symbol, records how the current layer was reached.
    \item The binary \emph{op-context} $o\in\{0,1\}$ indicates whether the current hidden state $h_t$ was produced by a plain layer or by an add operation.
\end{enumerate}
A single global policy is shared across these states (no
input-dependent context), giving a state space of size
$|\mathcal{S}| = L\cdot 3\cdot(2r{+}2)\cdot 2$.

\paragraph{Action Space}
At current layer \(\ell\), the policy chooses from three action families:
\begin{enumerate}
    \item \textbf{Local layer move:} apply layer \(\ell+\Delta\), where
    \(\Delta \in \{-r,\ldots,r\}\) and the destination layer is valid. This
    allows local skips, repeats, and backward revisits while biasing the search
    toward small perturbations of the standard forward pass.

    \item \textbf{Add-and-apply:} add a previous hidden state \(h_{t-b}\) to
    the current hidden state \(h_t\), using \(h_t + \gamma h_{t-b}\) with
    \(\gamma>0\), and then apply a local destination layer \(\ell+\Delta\).

    \item \textbf{\textsc{Rejoin}:} terminate the learned part of the route and
    execute the remaining standard suffix of transformer layers.
\end{enumerate}

At each routing step, we apply a feasibility mask that removes actions which would produce an invalid program, such as moving outside the layer range, exceeding the route-length budget, or adding a hidden state that has not yet been produced. The Markov policy is then renormalized over the remaining feasible actions, so sampling and Viterbi decoding always operate over valid routes. 
During decoding, a route is executed as an ordered sequence of frozen block
applications, and each program step gets its own attention cache slot. Thus a
revisited block \(f_\ell\) keeps a separate key/value history for each
application, so a route of \(T\) applied steps has the cache and decoding cost of
a \(T\)-block network with tied weights.

\paragraph{Policy Initialization}
The initial Markov policy is a forward-biased prior rather than a uniform
distribution over routes. We make the baseline route most probable because
accuracy-improving programs are expected to be slight modifications of the
standard forward pass rather than entirely random layer programs. The route starts at the first layer and the final
layer must terminate by rejoining the standard suffix:
\[
\rho(\ell)=\mathbf{1}\{\ell=0\},
\qquad
\pi_{\theta_0}(\textsc{Rejoin}\mid s)=1
\quad\text{if }\ell=L{-}1.
\]
For any earlier layer, let \(a_{\rightarrow}\) be the standard forward move
\(\ell\mapsto\ell+1\), let \(\mathcal{A}_{\mathrm{loc}}\) be the local layer
moves plus \textsc{Rejoin}, and let \(\mathcal{A}_{\mathrm{op}}\) be the
add-and-apply actions. Before feasibility masking, we set
\[
\begin{aligned}
\pi_{\theta_0}(a\mid s)
={}&
(1-\varepsilon_{\mathrm{loc}}-\varepsilon_{\mathrm{op}})
\mathbf{1}\{a=a_{\rightarrow}\}\\
&+
\frac{\varepsilon_{\mathrm{loc}}}{|\mathcal{A}_{\mathrm{loc}}|}
\mathbf{1}\{a\in\mathcal{A}_{\mathrm{loc}}\}\\
&+
\frac{\varepsilon_{\mathrm{op}}}{|\mathcal{A}_{\mathrm{op}}|}
\mathbf{1}\{a\in\mathcal{A}_{\mathrm{op}}\}.
\end{aligned}
\]
Thus \(a_{\rightarrow}\) receives the large remaining mass, while all local
deviations and add actions receive small exploration mass. The most likely
route under \(\pi_{\theta_0}\) is therefore the unmodified network,
\(\langle0,1,\dots,L{-}1\rangle\). This prior is also the default for states
that the search never visits: since the update uses pseudo-counts
\(\alpha\pi_{\theta_i}\), the empirical row \(\hat p_i\) and weighted visit
count \(m_i(s)\) give
\[
\hat{\pi}_{i+1}(\cdot\mid s)
=
\frac{\alpha}{\alpha+m_i(s)}\pi_{\theta_i}(\cdot\mid s)
+
\frac{m_i(s)}{\alpha+m_i(s)}\hat p_i(\cdot\mid s).
\]
Unvisited states have \(m_i(s)=0\) and keep the initialization exactly; visited
states move away from it only in proportion to the evidence collected for that
state.

\paragraph{Update Rule}
We update the Markov policy with an estimation-of-distribution procedure. At
iteration \(i\), we sample \(K\) candidate programs
\(\{r_j\}_{j=1}^{K}\) from the current policy \(\pi_{\theta_i}\), apply the
feasibility mask at each step, and evaluate each candidate on
\(\mathcal{S}_i\):
\[
q_j = \operatorname{Acc}_{\mathcal{S}_i}(r_j).
\]
The scores are converted into normalized update weights. Optionally, only the
top \(m\) candidates are kept as an elite set \(E_i\); otherwise
\(E_i=\{1,\ldots,K\}\):
\[
w_j =
\begin{cases}
\dfrac{\exp(\beta(q_j - \max_{k\in E_i} q_k))}
{\sum_{k\in E_i}\exp(\beta(q_k - \max_{\ell\in E_i} q_\ell))}
& \text{if } j \in E_i,\\[1.2em]
0 & \text{otherwise.}
\end{cases}
\]
Here \(\beta\) controls the sharpness of selection.

Let \(N_i(s,a)\) denote the weighted transition count for taking action \(a\)
from state \(s\). We initialize the counts with a smoothed copy of the current
transition table and then add the weighted transitions observed in the sampled
programs:
\[
N_i(s,a)
=
\alpha\,\pi_{\theta_i}(a\mid s)
+
\sum_{j=1}^{K}
w_j
\sum_{t}
\mathbf{1}\{s_{j,t}=s,\ a_{j,t}=a\}.
\]
The new policy is the normalized weighted maximum-likelihood estimate,
\[
\hat{\pi}_{i+1}(a\mid s)
=
\frac{N_i(s,a)}
{\sum_{a'} N_i(s,a')}.
\]
In the default update we set
\(\pi_{\theta_{i+1}}=\hat{\pi}_{i+1}\). For a smoother update, we optionally use
an exponential moving average,
\[
\pi_{\theta_{i+1}}
=
(1-\eta)\pi_{\theta_i}
+
\eta\hat{\pi}_{i+1},
\]
followed by row-wise normalization. This update increases the probability of
high-scoring route programs while retaining probability mass on nearby
alternatives through smoothing and feasibility-masked sampling.

\paragraph{Final Selection}
After the last update, the learned Markov policy defines a distribution over
valid route programs. Rather than selecting the best route seen during sampling,
we decode the most likely programs under the final policy using a top-\(k\)
Viterbi procedure. For a partial state \(s\), let \(\mathcal{A}(s)\) be the
set of feasible actions after applying the length and layer-bound masks. The
score of a complete program \(r=(a_1,\ldots,a_T)\) is its log-probability under
the masked policy,
\[
\log \pi_\theta(r)
=
\sum_{t=1}^{T}
\log \pi_\theta(a_t \mid s_t).
\]
The Viterbi recursion keeps the top \(k\) suffixes from each state:
\[
V_k(s)
=
\operatorname{TopK}_{a \in \mathcal{A}(s)}
\left[
\log \pi_\theta(a\mid s) + V_k(T(s,a))
\right],
\]
where \(T(s,a)\) is the next state after action \(a\). If \(a\) is
\textsc{Rejoin}, the recursion terminates by appending the standard remaining
suffix.

This produces a small candidate set \(\mathcal{C}_{\mathrm{vit}}\) of high
probability valid programs. We then evaluate these candidates on the validation
set and select
\[
\hat r
=
\operatorname*{arg\,max}_{r \in \mathcal{C}_{\mathrm{vit}}}
\operatorname{Acc}_{\mathcal{D}_{\mathrm{val}}}(r),
\]
breaking ties by higher training accuracy and then by shorter route length. The
test set is used only once, to report the final accuracy of the selected route
\(\hat r\).

\section{Experiments}

In our main experiments we will use the following baselines:

\begin{itemize}
    \item \textbf{Baseline:} The baseline follows the standard transformer execution path, applying every layer sequentially from the first layer to the last.


    \item \textbf{Dr.LLM} \citep{heakl2026drllm}: Dr.LLM trains lightweight
    per-layer routers from MCTS-derived route labels, providing a learned
    layer-routing baseline.

    \item \textbf{Dr.LLM+Ext.:} Dr.LLM+Ext. uses the same Dr.LLM-style router
    training procedure, but with the extended \ours{} action space.

    \item \textbf{\ours{} (top-1):} \ours{} (top-1) deploys the single most
    probable route under the learned policy, that is the rank-1 Viterbi
    candidate.
    
    \item \textbf{\ours{} (no Vit.):} \ours{} (no Vit.) removes decoding
    altogether and deploys the highest training-accuracy route the search ever
    sampled.

    \item \textbf{\ours{}:} \ours{} learns a task-specific Markov policy over
    feasible layer-routing programs, decodes top-\(k\) candidates with Viterbi,
    and selects a single validation-best route for test-time evaluation.
\end{itemize}

Additional details on dataset splits, generation settings, route-search
hyperparameters, and baseline training are provided in Appendix~\ref{app:experimental_details}.

\paragraph{Main Results} We compare \ours{} against baselines on mathematical
reasoning \textbf{GSM8K}~\citep{cobbe2021training} and
\textbf{MATH500}~\citep{lightman2024lets}, knowledge-intensive question answering
\textbf{OpenBookQA}~\citep{mihaylov-etal-2018-suit}, \textbf{MedQA}~\citep{jin2021disease},
\textbf{SciQ}~\citep{welbl2017crowdsourcing}, \textbf{MMLU-Pro}~\citep{wang2024mmlu},
\textbf{SVAMP}~\citep{patel2021nlp}, \textbf{ASDiv}~\citep{miao2020diverse},
\textbf{MAWPS}~\citep{koncel-kedziorski-etal-2016-mawps},
\textbf{GSM8K-Hard}~\citep{gao2023pal}, \textbf{GSM8K-Plus}~\citep{li2024gsm},
\textbf{multistep-arithmetic}, and \textbf{object counting}~\citep{suzgun2023challenging}. We
additionally report results on meta-llama/Llama-3.2-3B-Instruct in Appendix~\ref{app:llama}.
Table~\ref{tab:main_results} shows that \ours{} attains the best average
accuracy on every model except the reasoning-distilled one, and that its margin
over the unrouted baseline is widest exactly where that baseline is weakest,
narrowing to a smaller but consistent gain once the model is already strong.
Dr.\ LLM in its original skip-and-repeat action space frequently ends up below
the unrouted baseline. Dr.\ LLM+Ext. keeps the same routers, supervision, and
label budget and differs only in adopting the \ours{} action space, yet is
substantially stronger on every model we test, indicating that the action space
itself is a decisive part of what makes layer routing help rather than hurt.

\begin{table*}[!t]
\centering
\small
\setlength{\tabcolsep}{1.5pt}
\renewcommand{\arraystretch}{1.05}
\arrayrulecolor{SoftRule}

\resizebox{\textwidth}{!}{
\begin{tabular}{lccccccccccccc|c}
\toprule
\rowcolor{HeaderSoft}
\textcolor{HeaderText}{\textbf{Method}}
& \textcolor{HeaderText}{\textbf{GSM8K}}
& \textcolor{HeaderText}{\textbf{MATH500}}
& \textcolor{HeaderText}{\textbf{MedQA}}
& \textcolor{HeaderText}{\textbf{OpenBookQA}}
& \textcolor{HeaderText}{\textbf{SciQ}}
& \textcolor{HeaderText}{\textbf{MMLU-Pro}}
& \textcolor{HeaderText}{\textbf{SVAMP}}
& \textcolor{HeaderText}{\textbf{ASDiv}}
& \textcolor{HeaderText}{\textbf{MAWPS}}
& \textcolor{HeaderText}{\textbf{GSM8K-Hard}}
& \textcolor{HeaderText}{\textbf{GSM8K-Plus}}
& \textcolor{HeaderText}{\textbf{MS-Arith}}
& \textcolor{HeaderText}{\textbf{Obj-Count}}
& \textcolor{HeaderText}{\textbf{Avg.}} \\
\midrule

\modelbar{QwenBlue}{Qwen3-1.7B}
Baseline        & 42.84& 34.00& 38.88& \yellowc 68.00& 92.60& 22.13& 72.67& \orangec 90.25& \yellowc 57.88& 18.18& 29.10& 27.00& 61.00& 50.35 \\
Dr.LLM          & 50.04\std{4.42}& 20.33\std{3.06}& 35.35\std{3.45}& 54.93\std{5.59}& 79.53\std{2.44}& 17.88\std{1.75}& 55.67\std{6.98}& \yellowc 72.64\std{12.78}& 53.91\std{4.62}& 15.67\std{1.96}& 26.20\std{2.46}& 74.07\std{4.98}& 43.00\std{1.00}& 46.09 \\
Dr.LLM+Ext.     & 42.71\std{0.68}& \yellowc 35.33\std{1.53}& \yellowc 39.15\std{0.87}& 56.67\std{10.71}& 77.67\std{24.31}& 23.12\std{0.38}& 70.22\std{1.35}& \orangec 90.25\std{1.37}& 56.35\std{1.15}& 17.66\std{0.48}& \yellowc 30.70\std{0.26}& \redc 96.48\std{0.32}& \yellowc 64.00\std{2.65}& 53.87 \\
\ours{} (no Vit.) & \orangec 62.62\std{11.73}& \orangec 36.33\std{4.16}& \redc 44.70\std{2.93}& \redc 78.20\std{0.57}& \yellowc 92.93\std{0.12}& \redc 30.00\std{2.47}& \redc 83.78\std{4.14}& \redc 96.12\std{1.01}& \redc 65.64\std{2.85}& \redc 30.93\std{5.40}& \orangec 47.30\std{10.06}& \orangec 86.67\std{7.51}& \orangec 68.33\std{3.21}& \redc 63.35 \\
\ours{} (top-1) & \yellowc 62.37\std{11.42}& \orangec 36.33\std{4.16}& \redc 44.70\std{2.93}& \redc 78.20\std{0.57}& \orangec 92.97\std{0.15}& \yellowc 26.50\std{5.66}& \orangec 83.67\std{4.33}& \redc 96.12\std{1.01}& \redc 65.64\std{2.85}& \yellowc 28.53\std{3.26}& \orangec 47.30\std{10.06}& \yellowc 84.33\std{6.51}& \redc 69.33\std{4.93}& \yellowc 62.77 \\
\textbf{\ours{}}& \redc 63.56\std{12.15}& \redc 38.00\std{7.00}& \orangec 42.84\std{2.91}& \orangec 77.80\std{3.96}& \redc 93.03\std{0.25}& \orangec 29.38\std{1.59}& \yellowc 83.11\std{3.75}& \redc 96.12\std{1.01}& \orangec 64.17\std{1.97}& \orangec 29.99\std{5.03}& \redc 47.70\std{9.37}& \yellowc 84.33\std{6.51}& \redc 69.33\std{4.93}& \orangec \textbf{63.03} \\
\midrule

\modelbar{QwenTeal}{Qwen/Qwen3-4B}
Baseline        & 74.75& \orangec 36.67& \redc 62.48& \redc 90.47& \redc 95.73& 44.42& 87.44& \yellowc 98.95& \yellowc 72.12& 32.39& 57.33& \redc 100.00& \yellowc 91.33& 72.62 \\
Dr.LLM          & 70.86\std{3.62}& 31.67\std{2.31}& 58.18\std{0.65}& \yellowc 85.80\std{3.27}& 93.47\std{0.72}& 41.12\std{2.09}& 83.00\std{0.33}& 85.74\std{9.69}& 68.27\std{2.69}& 29.15\std{1.63}& 48.07\std{4.45}& 89.07\std{3.35}& 74.33\std{5.13}& 66.06 \\
Dr.LLM+Ext.     & \redc 79.18\std{3.31}& 35.33\std{2.08}& \orangec 62.19\std{0.20}& \orangec 90.13\std{1.17}& 90.73\std{6.88}& 43.67\std{0.31}& 84.56\std{1.64}& \redc 99.27\std{0.48}& \yellowc 72.12\std{0.19}& 30.83\std{1.73}& 54.67\std{2.40}& \yellowc 99.07\std{1.16}& \orangec 91.67\std{2.52}& 71.80 \\
\ours{} (no Vit.) & 75.03\std{0.50}& 35.67\std{1.53}& 61.38\std{0.36}& \redc 90.47\std{0.64}& \yellowc 94.67\std{0.90}& \orangec 45.33\std{0.52}& \yellowc 88.00\std{1.67}& \yellowc 98.95\std{0.48}& \orangec 72.56\std{0.40}& \orangec 37.51\std{5.65}& \yellowc 60.50\std{3.50}& \orangec 99.26\std{0.32}& \redc 94.33\std{1.15}& \yellowc 73.36 \\
\ours{} (top-1) & \yellowc 76.12\std{2.24}& \yellowc 36.33\std{2.52}& \yellowc 61.64\std{0.87}& \redc 90.47\std{0.64}& \orangec 95.70\std{0.10}& \yellowc 44.83\std{0.95}& \orangec 88.78\std{1.64}& \redc 99.27\std{0.48}& \orangec 72.56\std{0.40}& \redc 37.72\std{5.29}& \orangec 60.60\std{3.36}& \orangec 99.26\std{0.32}& \redc 94.33\std{1.15}& \orangec 73.66 \\
\textbf{\ours{}}& \orangec 77.99\std{2.91}& \redc 37.00\std{1.00}& \redc 62.48\std{0.37}& \redc 90.47\std{0.31}& \redc 95.73\std{0.15}& \redc 46.04\std{1.13}& \redc 89.67\std{1.20}& \orangec 99.06\std{0.54}& \redc 74.36\std{3.07}& \yellowc 36.47\std{4.35}& \redc 62.70\std{1.28}& \redc 100.00\std{0.00}& \orangec 91.67\std{2.31}& \redc \textbf{74.12} \\
\midrule

\modelbar{QwenOrange}{Qwen/Qwen3-8B}
Baseline        & 75.39& \orangec 41.33& \yellowc 68.08& \yellowc 93.87& \orangec 96.17& \orangec 50.67& 89.78& \redc 99.48& 73.91& \yellowc 38.14& 59.33& \yellowc 98.89& \redc 97.67& \orangec 75.59 \\
Dr.LLM          & 75.84\std{1.05}& 28.67\std{7.57}& 66.22\std{1.47}& 90.53\std{1.42}& 95.00\std{0.50}& 46.92\std{0.94}& 82.67\std{6.39}& 95.28\std{2.37}& 69.49\std{5.02}& 31.56\std{0.18}& 51.10\std{1.31}& 94.63\std{2.80}& 77.33\std{5.03}& 69.63 \\
Dr.LLM+Ext.     & \redc 78.85\std{0.62}& \yellowc 40.33\std{2.08}& \redc 68.34\std{0.52}& 93.27\std{0.70}& 95.40\std{1.51}& 49.54\std{0.07}& \orangec 90.22\std{0.69}& \orangec 99.16\std{0.18}& \yellowc 74.68\std{0.40}& 37.72\std{0.79}& 58.57\std{0.85}& \redc 99.63\std{0.64}& \orangec 96.00\std{1.73}& \yellowc 75.52 \\
\ours{} (no Vit.) & 76.32\std{0.99}& 38.33\std{1.53}& 67.22\std{1.88}& 93.67\std{2.00}& 95.37\std{1.27}& 49.25\std{1.87}& \yellowc 90.00\std{0.33}& \yellowc 99.06\std{0.31}& \redc 76.35\std{2.40}& \orangec 38.45\std{1.45}& \redc 61.00\std{1.25}& 98.33\std{0.56}& \yellowc 95.33\std{1.15}& 75.28 \\
\ours{} (top-1) & \yellowc 76.57\std{0.66}& 38.33\std{1.53}& 67.45\std{2.11}& \redc 94.47\std{0.95}& \yellowc 96.07\std{0.15}& \yellowc 50.17\std{0.59}& \orangec 90.22\std{0.51}& \yellowc 99.06\std{0.31}& \redc 76.35\std{2.40}& 37.72\std{1.10}& \yellowc 60.03\std{1.17}& 98.70\std{0.32}& \yellowc 95.33\std{0.58}& 75.42 \\
\textbf{\ours{}}& \orangec 77.71\std{2.63}& \redc 41.67\std{0.58}& \orangec 68.29\std{0.24}& \orangec 94.20\std{0.92}& \redc 96.33\std{0.15}& \redc 50.79\std{0.36}& \redc 90.33\std{0.33}& \redc 99.48\std{0.48}& \orangec 76.15\std{1.07}& \redc 39.92\std{0.72}& \orangec 60.97\std{1.04}& \orangec 99.07\std{0.32}& \redc 97.67\std{0.58}& \redc \textbf{76.35} \\
\midrule

\modelbar{QwenRed}{Qwen/Qwen3-14B (4-bit)}
Baseline        & 83.93& \orangec 42.67& 73.74& \yellowc 95.13& \orangec 97.07& \redc 56.42& \yellowc 91.67& \yellowc 99.16& \orangec 75.06& \yellowc 42.11& 63.77& \redc 100.00& \redc 100.00& \yellowc 78.52 \\
Dr.LLM          & 81.50\std{1.18}& 36.33\std{3.79}& 71.62\std{1.56}& 93.93\std{1.42}& 95.37\std{0.35}& 54.12\std{1.39}& 86.00\std{1.76}& 96.33\std{1.61}& 70.77\std{3.97}& 37.30\std{2.99}& 56.23\std{1.10}& \yellowc 96.85\std{0.85}& 85.67\std{1.53}& 74.00 \\
Dr.LLM+Ext.     & \yellowc 84.86\std{0.39}& 41.67\std{1.15}& 63.68\std{17.94}& \orangec 95.47\std{0.12}& 96.80\std{0.10}& 55.12\std{0.88}& 90.78\std{1.64}& \yellowc 99.16\std{0.18}& 74.42\std{0.33}& \yellowc 42.11\std{1.31}& 63.67\std{0.12}& \orangec 99.26\std{1.28}& 94.33\std{6.35}& 77.03 \\
\ours{} (no Vit.) & \redc 86.38\std{2.11}& \yellowc 42.33\std{4.04}& \yellowc 73.95\std{0.05}& 95.07\std{1.67}& 96.97\std{0.23}& \yellowc 55.46\std{0.19}& \redc 92.11\std{0.51}& \orangec 99.27\std{0.18}& \yellowc 74.74\std{1.83}& 41.48\std{2.04}& \orangec 65.93\std{0.65}& \redc 100.00\std{0.00}& \yellowc 99.00\std{1.00}& \orangec 78.67 \\
\ours{} (top-1) & 83.83\std{0.24}& 41.67\std{1.53}& \orangec 74.10\std{0.71}& 95.00\std{0.20}& \yellowc 97.03\std{0.12}& \orangec 56.17\std{1.09}& 91.11\std{0.77}& \yellowc 99.16\std{0.36}& 74.36\std{0.11}& \orangec 42.53\std{0.96}& \yellowc 64.23\std{0.51}& \redc 100.00\std{0.00}& \orangec 99.67\std{0.58}& 78.37 \\
\textbf{\ours{}}& \orangec 86.20\std{1.51}& \redc 44.67\std{2.08}& \redc 74.37\std{0.45}& \redc 95.53\std{0.95}& \redc 97.13\std{0.06}& \redc 56.42\std{0.26}& \orangec 92.00\std{0.58}& \redc 99.48\std{0.18}& \redc 75.71\std{0.87}& \redc 42.74\std{0.48}& \redc 66.57\std{0.76}& \redc 100.00\std{0.00}& \redc 100.00\std{0.00}& \redc \textbf{79.29} \\
\midrule

\modelbar{MistralRose}{mistralai/Mixtral-8x7B-Instruct-v0.1 (4-bit)}
Baseline        & 67.15& 20.33& \orangec 55.73& 75.00& 90.10& \yellowc 24.21& 76.67& 91.93& \yellowc 64.68& 32.60& \yellowc 49.27& \yellowc 93.33& 66.33& 62.10 \\
Dr.LLM          & 65.20\std{0.93}& 17.67\std{0.58}& 51.66\std{0.70}& 69.93\std{3.80}& 88.90\std{0.26}& 21.29\std{0.63}& 66.22\std{1.26}& 80.82\std{1.63}& 54.55\std{2.25}& 29.26\std{1.73}& 38.20\std{1.47}& 78.33\std{0.96}& 47.00\std{1.73}& 54.54 \\
Dr.LLM+Ext.     & \redc 68.76\std{1.38}& \redc 25.00\std{0.00}& \yellowc 55.70\std{0.00}& \yellowc 76.20\std{0.00}& 89.80\std{0.00}& 22.50\std{0.00}& \yellowc 77.33\std{0.58}& 90.88\std{0.00}& 63.97\std{0.22}& \orangec 33.75\std{0.18}& 49.00\std{0.36}& \yellowc 93.33\std{0.00}& 65.33\std{0.58}& 62.43 \\
\ours{} (no Vit.) & \yellowc 67.32\std{1.05}& \yellowc 22.67\std{3.51}& \redc 56.52\std{0.94}& \redc 77.70\std{1.56}& \redc 90.85\std{0.35}& \orangec 24.62\std{1.84}& \orangec 77.89\std{0.77}& \yellowc 92.24\std{1.58}& \orangec 65.32\std{2.13}& 32.29& \redc 50.83\std{1.10}& \orangec 94.63\std{1.79}& \yellowc 67.00\std{4.36}& \orangec 63.07 \\
\ours{} (top-1) & 67.22\std{0.16}& 21.33\std{4.04}& 55.07\std{1.11}& 74.67\std{1.22}& \yellowc 90.40\std{0.61}& \yellowc 24.21\std{0.07}& \redc 78.00\std{0.88}& \orangec 92.56\std{1.10}& 64.49\std{0.11}& \yellowc 33.39\std{1.55}& \orangec 49.53\std{1.46}& \yellowc 93.33\std{1.47}& \orangec 67.67\std{2.52}& \yellowc 62.45 \\
\textbf{\ours{}}& \orangec 67.50\std{0.92}& \orangec 24.33\std{2.31}& \redc 56.52\std{0.94}& \orangec 77.33\std{2.58}& \orangec 90.67\std{0.21}& \redc 25.12\std{0.76}& \yellowc 77.33\std{1.53}& \redc 93.40\std{2.27}& \redc 65.77\std{1.45}& \redc 34.33\std{0.22}& \orangec 49.53\std{1.46}& \redc 94.81\std{0.85}& \redc 68.00\std{2.00}& \redc \textbf{63.43} \\
\midrule

\modelbar{DeepSeekCyan}{deepseek-ai/DeepSeek-R1-Distill-Llama-8B}
Baseline        & 20.67& 31.00& \redc 48.10& 80.60& \redc 90.57& \yellowc 32.62& 33.78& 45.81& 35.58& \yellowc 20.79& \yellowc 46.47& 36.67& 34.67& 42.87 \\
Dr.LLM          & \orangec 60.48\std{0.53}& 33.33\std{3.21}& 46.61\std{1.41}& 76.60\std{1.74}& 87.13\std{1.16}& 30.79\std{1.78}& \orangec 67.78\std{3.15}& \yellowc 84.28\std{1.96}& \orangec 60.96\std{1.85}& 20.38\std{0.83}& 39.23\std{3.32}& 81.11\std{1.47}& \orangec 73.33\std{6.11}& \orangec 58.62 \\
Dr.LLM+Ext.     & \redc 61.36\std{0.27}& \redc 40.00\std{1.73}& \yellowc 47.97\std{0.92}& 79.67\std{1.10}& \yellowc 88.70\std{1.40}& 32.42\std{1.13}& \redc 74.89\std{0.19}& \redc 90.67\std{0.18}& \redc 64.04\std{0.00}& \redc 21.63\std{0.31}& 46.17\std{0.65}& \redc 90.00\std{0.00}& \redc 75.33\std{7.23}& \redc 62.53 \\
\ours{} (no Vit.) & \yellowc 31.19\std{5.58}& \orangec 34.33\std{3.06}& \orangec 48.00\std{0.00}& \redc 81.30\std{0.14}& \orangec 90.20\std{0.28}& \orangec 33.06\std{0.62}& \yellowc 58.00\std{2.40}& \orangec 86.06\std{7.42}& 54.29\std{1.49}& 20.53\std{0.22}& 45.20\std{1.84}& \yellowc 82.41\std{4.17}& \yellowc 69.67\std{12.34}& \yellowc 56.48 \\
\ours{} (top-1) & 20.34\std{0.23}& \yellowc 33.67\std{2.89}& 47.92\std{0.11}& \yellowc 81.00\std{0.57}& \orangec 90.20\std{0.28}& \orangec 33.06\std{0.62}& 33.22\std{1.26}& 45.28\std{1.44}& 35.45\std{0.68}& 20.22\std{0.66}& \orangec 46.70\std{0.57}& 36.48\std{1.16}& 34.67\std{1.53}& 42.94 \\
\textbf{\ours{}}& 30.53\std{5.62}& 33.00\std{3.61}& \redc 48.10\std{0.73}& \orangec 81.27\std{0.58}& \redc 90.57\std{0.51}& \redc 33.25\std{0.54}& \yellowc 58.00\std{2.40}& 81.66\std{4.52}& \yellowc 56.03\std{2.32}& \orangec 21.11\std{0.90}& \redc 46.83\std{0.38}& \orangec 83.15\std{4.32}& 63.33\std{7.02}& \textbf{55.91} \\
    \bottomrule
\end{tabular}
}
\caption{Performance comparison across prompting and optimization methods. Colored bars denote different model families. Within each model and benchmark column, cells are shaded by rank across methods (red: best, orange: second, yellow: third); ties share a color. Values with gray subscripts report mean and standard deviation over 3 seeds.}
\label{tab:main_results}
\end{table*}

\paragraph{\ours{} Without Viterbi or Validation Selection}
\ours{} (no Vit.) keeps the learned policy but discards decoding, deploying the
highest training-accuracy route the search happened to sample.
\ours{} (top-1) instead keeps decoding but discards validation selection,
deploying the single most probable route under the final policy.
Full \ours{} is the strongest of the three in Table~\ref{tab:main_results}, best
on four of the six models and on the average over them .
Both stages appear to contribute, and each replaces a lucky draw with a more
principled choice: Viterbi reconstructs candidates from the transition table
rather than trusting whichever routes the sampler happened to visit, and
validation then ranks those candidates on held-out data rather than on the
training score the search has already optimized.

\paragraph{Statistical Significance}
Seed replicates of one benchmark share a test set and a deterministic baseline
run, so they are not independent observations. We therefore first average the
accuracy difference
$d=\operatorname{Acc}_{\ours}-\operatorname{Acc}_{\mathrm{base}}$ over seeds
within each benchmark, and treat the $13$ resulting per-benchmark differences as
the units of analysis. Per model we run a one-sided Wilcoxon signed-rank test of
$H_0$: \ours{} does not improve over the baseline (no positive shift in $d$)
against $H_1$: \ours{} improves accuracy. Table~\ref{tab:significance}
reports the mean improvement $\bar d$, which equals the per-model average gain
in Table~\ref{tab:main_results}, a 95\% confidence interval from a cluster
bootstrap over benchmarks ($10^{5}$ resamples), the counts of benchmarks
improved/tied/degraded, and $p$-values Holm-corrected across the six models.
Every model improves significantly, degrades on no benchmark, and excludes zero.

\begin{table}[!tb]
\centering
\small
\setlength{\tabcolsep}{5.5pt}
\renewcommand{\arraystretch}{1.18}
\arrayrulecolor{SoftRule}
\resizebox{\columnwidth}{!}{
\begin{tabular}{lcccc}
\toprule
\rowcolor{HeaderSoft}
\textcolor{HeaderText}{\textbf{Model}}
& \textcolor{HeaderText}{\textbf{$\bar d$ (pts)}}
& \textcolor{HeaderText}{\textbf{95\% CI}}
& \textcolor{HeaderText}{\textbf{W/T/L}}
& \textcolor{HeaderText}{\textbf{$p_{\mathrm{Holm}}$}} \\
\midrule
Qwen3-1.7B                   & $+12.68$ & $[\,6.70,\ 21.38\,]$ & 13/0/0 & $7.3{\times}10^{-4}$ \\
Qwen3-4B                     & $+1.50$  & $[\,0.62,\ 2.51\,]$  & 9/4/0  & $2.0{\times}10^{-3}$ \\
Qwen3-8B                     & $+0.76$  & $[\,0.33,\ 1.25\,]$  & 11/2/0 & $2.0{\times}10^{-3}$ \\
Qwen3-14B (4-bit)            & $+0.78$  & $[\,0.33,\ 1.31\,]$  & 10/3/0 & $2.0{\times}10^{-3}$ \\
Mixtral-8x7B (4-bit)         & $+1.33$  & $[\,0.86,\ 1.91\,]$  & 13/0/0 & $7.3{\times}10^{-4}$ \\
DeepSeek-R1-Distill-Llama-8B & $+13.04$ & $[\,5.07,\ 22.00\,]$ & 11/2/0 & $2.0{\times}10^{-3}$ \\
\bottomrule
\end{tabular}
}
\caption{Statistical significance of \ours{} over the baseline, with benchmarks
as the unit of analysis.}
\label{tab:significance}
\end{table}

\paragraph{State-Space Ablation}
We ablate the context used by the Markov transition model on Qwen3-1.7B by removing one state factor at a time: the budget phase, the incoming layer delta, or the operator context. The full state performs best overall, with the largest drop coming from removing the budget-phase factor, indicating that routing decisions depend strongly on where the program sits in its computation budget.

\begin{table}[!tb]
\centering
\small
\setlength{\tabcolsep}{5.5pt}
\renewcommand{\arraystretch}{1.18}
\arrayrulecolor{SoftRule}
\resizebox{\columnwidth}{!}{
\begin{tabular}{lcccrr}
\toprule
\rowcolor{HeaderSoft}
\textcolor{HeaderText}{\textbf{State Variant}}
& \textcolor{HeaderText}{\textbf{Phase}}
& \textcolor{HeaderText}{\textbf{Delta}}
& \textcolor{HeaderText}{\textbf{Op Ctx.}}
& \textcolor{HeaderText}{\textbf{Avg.}}
& \textcolor{HeaderText}{\textbf{Drop}} \\
\midrule
\rowcolor{QwenBlue!8}
\textbf{Full state} & \kept & \kept & \kept & \textbf{63.03} & -- \\
No delta            & \kept & \dropped & \kept & 62.54\std{3.32} & -0.49 \\
No op context       & \kept & \kept & \dropped & 61.48\std{3.59} & -1.55 \\
No phase            & \dropped & \kept & \kept & 59.56\std{5.19} & -3.47 \\
\bottomrule
\end{tabular}
}
\caption{State-space ablation on Qwen3-1.7B. Averages are over all the 13 benchmarks used in Table~\ref{tab:main_results}.}
\label{tab:state_space_ablation}
\end{table}

\begin{figure}[!tb]
\centering
\includegraphics[width=\columnwidth]{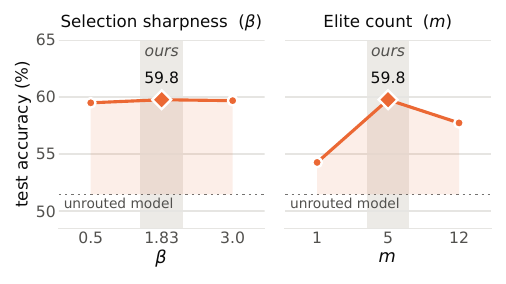}
\caption{Accuracy on Qwen3-1.7B when sweeping the selection
sharpness $\beta$ (left) and the elite count $m$ (right) one at a time around
our configuration (shaded, diamond).}
\label{fig:hparam}
\end{figure}

\paragraph{Hyperparameter Analysis}
We sweep one search hyperparameter at a time around our configuration on
Qwen3-1.7B: the selection sharpness $\beta$ and the elite count $m$
(Figure~\ref{fig:hparam}). Accuracy is essentially flat in $\beta$, since the
elite set already applies a hard filter before the update and leaves the softer
reweighting little to decide. The elite count matters more, and our setting sits
between two failure modes: $m=1$ fits the transition counts to a single
trajectory and commits to it before the evidence supports doing so, making it
the only setting significantly worse than ours, while $m=12$ keeps too broad a
range of candidates for the update to stay selective and dilutes the counts with
noise. Every setting still improves over the unrouted model, so \ours{} does not
depend on a narrow hyperparameter choice.

\begin{figure}[!tb]
\centering
\includegraphics[width=\columnwidth]{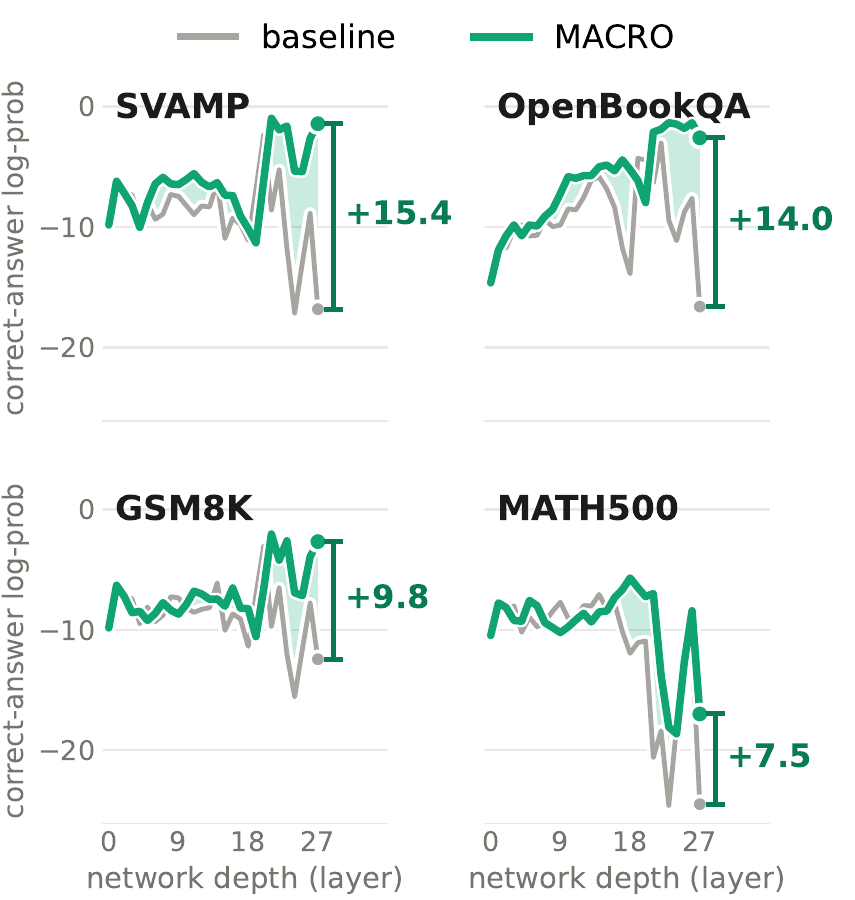}
\caption{Logit-lens
log-probability of the correct answer at each layer of Qwen3-1.7B for baseline route and \ours{} route}
\label{fig:mech_interp}
\end{figure}

\paragraph{Mechanistic Interpretability of \ours{}}
To understand \emph{how} a route improves accuracy without any weight update, we
apply a logit lens: at each layer we read the running hidden state through the
model's own output projection and track the probability assigned to the correct
answer token. Figure~\ref{fig:mech_interp} shows that the standard forward pass
can represent the answer internally and then suppress it near the output, while
\ours{} preserves this signal through the final readout. Because the model's own
generated reasoning is held fixed in context, this suggests that routing improves
how an already-computed answer is exposed rather than changing the reasoning
itself.

\begin{table}[!t]
\centering
\small
\setlength{\tabcolsep}{2.4pt}
\renewcommand{\arraystretch}{1.18}
\arrayrulecolor{SoftRule}
\begin{tabular}{l:ccc:ccc}
\toprule
\rowcolor{HeaderSoft}
\textcolor{HeaderText}{\textbf{Model}}
& \multicolumn{3}{c:}{\textcolor{HeaderText}{\textbf{In-Domain}}}
& \multicolumn{3}{c}{\textcolor{HeaderText}{\textbf{Out-of-Domain}}} \\
\rowcolor{HeaderSoft}
& \textcolor{HeaderText}{\textbf{Base.}}
& \textcolor{HeaderText}{\textbf{\ours{}}}
& \textcolor{HeaderText}{\textbf{$\Delta$}}
& \textcolor{HeaderText}{\textbf{Base.}}
& \textcolor{HeaderText}{\textbf{\ours{}}}
& \textcolor{HeaderText}{\textbf{$\Delta$}} \\
\midrule
Qwen3-1.7B & 71.50 & 74.29 & $+2.79$ & 30.98 & 40.00 & $+9.02$ \\
Qwen3-4B   & 83.77 & 84.07 & $+0.30$ & 56.09 & 57.21 & $+1.12$ \\
\bottomrule
\end{tabular}
\caption{Results for the \emph{one model = one route} experiment. A single
shared route is trained on the combined training pool from the eight in-domain
benchmarks and evaluated on both the held-out test splits of those in-domain
benchmarks and the five out-of-domain benchmarks.}
\label{tab:one_model_one_route}
\end{table}

\paragraph{One Model, One Route}
To test whether a single shared route can transfer across tasks, we train one route policy on a combined training pool built from MAWPS, ASDiv, SVAMP, multistep-arithmetic, object counting, MATH500, OpenBookQA, and SciQ. We then evaluate the resulting route on the held-out test splits of those eight in-domain benchmarks and on the five out-of-domain benchmarks GSM8K, GSM8K-Hard, GSM8K-Plus, MMLU-Pro, and MedQA. Results are summarized in Table~\ref{tab:one_model_one_route}. For each model, we report the mean accuracy of the baseline forward pass and the mean accuracy of the routed candidate selected by the shared route, along with the aggregate in-domain and out-of-domain averages. Accuracies for all benchmarks and representative non-standard route strings for the two
Qwen models are provided in Appendix~\ref{app:best_routes}.

\begin{figure}[!tb]
\centering
\includegraphics[width=\columnwidth]{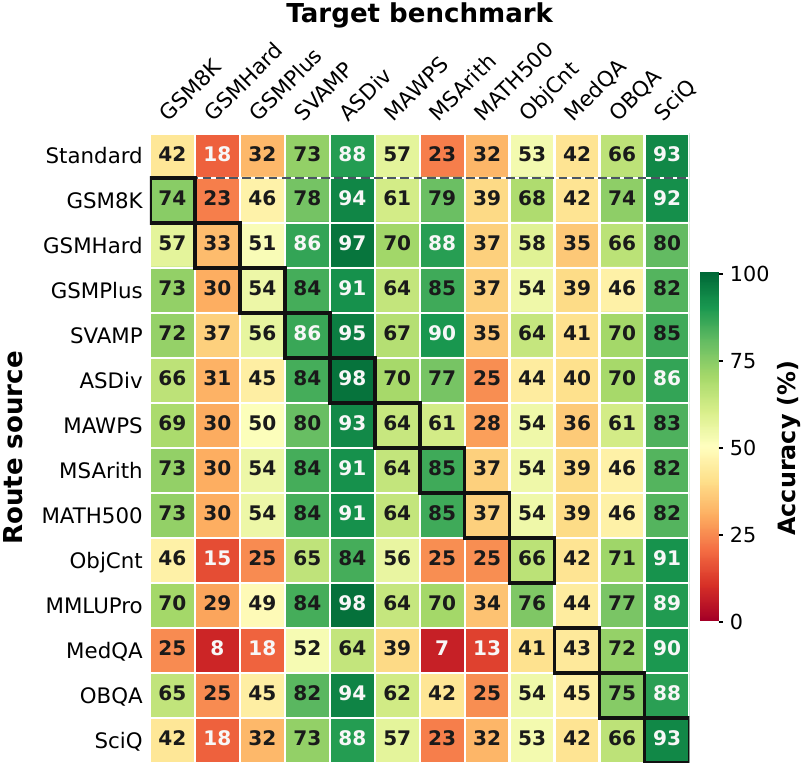}
\caption{Route-transfer heatmap on Qwen3-1.7B, with red-to-green shading from
lower to higher target accuracy. }
\label{tab:route_transfer}
\end{figure}

\paragraph{Route Transfer Across Benchmarks}
To test whether routes capture reusable computation or only benchmark-specific
quirks, we evaluate each Qwen3-1.7B route, unchanged, on every held-out target
benchmark. Only the route transfers: the prompts, scoring, data, and baseline
always come from the target benchmark. Figure~\ref{tab:route_transfer} shows this
accuracy matrix, where entry $(A,B)$ is the route searched on benchmark $A$
evaluated on target $B$. The highest-accuracy cells are mostly on the diagonal,
which is expected because each route is optimized for its home benchmark.
Off-diagonal transfer still appears, especially among related
mathematical-reasoning tasks, suggesting that some routes encode reusable model
computations rather than only task-specific effects.

\section{Conclusion}
We have introduced a lightweight dynamic routing method that outperforms the unrouted base model and Dr.LLM, the leading dynamic routing method across a plethora of benchmarks and LLM models. 
It is an interesting further research question why routing improves performance: Is it because routing improves capabilities or because an internally already present answer is better elicited? We argue that some evidence points towards the latter, including our limited mechanistic interpretability study.
We argue that routing should be taken much further: we want models that profit from much more extensive input-depending dynamic routing, allowing for another dimension of flexible test-time scaling.
\newpage

\bibliography{aaai2027}

\appendix

\setcounter{secnumdepth}{2}
\numberwithin{table}{section}
\numberwithin{figure}{section}

\newlength{\tmpw}
\newlength{\bmw}\settowidth{\bmw}{\small OpenBookQA}%
  \settowidth{\tmpw}{\small\textbf{Benchmark}}\ifdim\tmpw>\bmw\setlength{\bmw}{\tmpw}\fi
  \settowidth{\tmpw}{\small GSM8K-Plus}\ifdim\tmpw>\bmw\setlength{\bmw}{\tmpw}\fi
  \addtolength{\bmw}{6pt}
\newlength{\numw}\settowidth{\numw}{\small 100.00}%
  \settowidth{\tmpw}{\small\textbf{Baseline}}\ifdim\tmpw>\numw\setlength{\numw}{\tmpw}\fi
  \settowidth{\tmpw}{\small\textbf{\ours{}}}\ifdim\tmpw>\numw\setlength{\numw}{\tmpw}\fi
  \addtolength{\numw}{6pt}
\newlength{\delw}\settowidth{\delw}{\small +40.57}%
  \settowidth{\tmpw}{\small\textbf{$\Delta$}}\ifdim\tmpw>\delw\setlength{\delw}{\tmpw}\fi
  \addtolength{\delw}{6pt}
\newlength{\layw}\settowidth{\layw}{\small 40$\to$46}%
  \settowidth{\tmpw}{\small\textbf{Layers}}\ifdim\tmpw>\layw\setlength{\layw}{\tmpw}\fi
  \addtolength{\layw}{6pt}
\newlength{\routew}\settowidth{\routew}{\small\ttfamily L0-L18 L18-L22 L24(+h-6) L25-L27 L27-L31 RJ}%
  \addtolength{\routew}{6pt}

\section{Additional Experimental Results on Llama}
\label{app:llama}

We additionally evaluate \ours{} on meta-llama/Llama-3.2-3B-Instruct. For this
model we report the standard forward-pass baseline and \ours{} across all
benchmarks in Table~\ref{tab:llama_results}. The results follow the same protocol and rank-based color scheme as the
main results table. Consistent with the other
models, \ours{} matches or improves over the baseline on every benchmark;
pooling the per-dataset, per-seed accuracy differences, the mean improvement is
$\bar d = +0.35$ points (95\% bootstrap CI $[0.03,\ 0.79]$; one-sided Wilcoxon
signed-rank $p = 3.3\times10^{-2}$).

\begin{table*}[tp]
\centering
\small
\setlength{\tabcolsep}{5.0pt}
\renewcommand{\arraystretch}{1.12}
\arrayrulecolor{black}
\resizebox{\textwidth}{!}{
\begin{tabular}{lcccccccccccccc}
\toprule
\textbf{Method}
& \textbf{GSM8K}
& \textbf{MATH500}
& \textbf{MedQA}
& \textbf{OpenBookQA}
& \textbf{SciQ}
& \textbf{MMLU-Pro}
& \textbf{SVAMP}
& \textbf{ASDiv}
& \textbf{MAWPS}
& \textbf{GSM8K-Hard}
& \textbf{GSM8K-Plus}
& \textbf{MS-Arith}
& \textbf{Obj-Count}
& \textbf{Avg.} \\
\midrule
\multicolumn{15}{l}{\textbf{Model: meta-llama/Llama-3.2-3B-Instruct}} \\
Baseline        & 75.26& 42.33& 57.84& 77.53& 88.03& 33.21& 83.83& 96.12& 63.97& 26.33& 58.07& 97.41& 56.33& 65.87 \\
\textbf{\ours{}}& 75.26\std{0.00}& 42.33\std{0.00}& 57.84\std{0.00}& 77.53\std{0.31}& 88.40\std{0.35}& 33.21\std{0.00}& 83.83\std{0.00}& 96.12\std{0.00}& 63.97\std{0.00}& 26.33\std{0.00}& 58.07\std{0.00}& 97.41\std{0.00}& 60.33\std{2.31}& \textbf{66.20} \\
\bottomrule
\end{tabular}
}
\caption{Additional results on meta-llama/Llama-3.2-3B-Instruct.}
\label{tab:llama_results}
\end{table*}

\section{Best Discovered Routes}
\label{app:best_routes}

Tables~\ref{tab:best_routes_a} through~\ref{tab:best_routes_c} list, for every model and
benchmark, the chosen route. Routes built only
from local moves dominate on the larger models, while the small and distilled
models more often select routes that also contain an add operation.

\paragraph{Route notation}
Each \textbf{Best route} entry is read left to right as the sequence of
transformer layers applied to the running hidden state. \texttt{L$i$} applies
layer $i$ as a plain (identity) step, and \texttt{L$i$-L$j$} abbreviates the
consecutive run \texttt{L$i$\,L$i{+}1$\,$\cdots$\,L$j$}; a segment whose start
index drops below the preceding layer marks a backward revisit or repeat. A
parenthesized suffix denotes an add-and-apply step: \texttt{L$\ell$(+h-$b$)}
forms $h_t + \gamma\,h_{t-b}$ from the hidden state $b$ produced-states back and
then applies layer $\ell$, with the default coefficient $\gamma=1.0$ left
implicit; the subtract variant carries an explicit sign and coefficient, e.g.\
\texttt{L2(-0.25h-1)} applies $h_t - 0.25\,h_{t-1}$. \texttt{RJ} is
\textsc{Rejoin}: it terminates the learned prefix and appends the remaining
standard forward layers up to the last layer, so every route ends in
\texttt{RJ}. Rows marked \emph{standard forward pass} indicate that the raw
candidate is semantically equivalent to the unmodified baseline route after
\textsc{Rejoin} expansion, e.g.\ \texttt{L0 RJ} or \texttt{L0-L2 RJ}. The \textbf{Layers} column reports the
layer-application count (KV-cache depth) of the baseline and of the listed route
as \texttt{baseline$\to$route}, and \textbf{$\Delta$} is the listed \ours{} test
accuracy minus the baseline.

\paragraph{Representative non-standard routes for the one-model-one-route experiment}
Table~\ref{tab:one_model_one_route_benchmarks} reports, for each of the Qwen3-1.7B
and Qwen3-4B models, accuracy averaged over 3 seeds of the single shared policy
trained over the combined pool of in-domain benchmarks, evaluated on test splits
of both in-domain and out-of-domain benchmarks. The layer range and route printed below the table are from one representative seed, given as an illustrative example of the kind of routing
the policy discovers.

\newcommand{\modelbarroute}[1]{\rowcolor{ModelBarGray}\multicolumn{6}{l}{\textcolor{white}{\textbf{Model: #1}}} \\}

\begin{table*}[tp]
\centering
\small
\setlength{\tabcolsep}{5.0pt}
\renewcommand{\arraystretch}{1.10}
\arrayrulecolor{SoftRule}
\resizebox{\textwidth}{!}{
\begin{tabular}{p{\bmw}p{\numw}p{\numw}p{\delw}p{\layw}p{\routew}}
\toprule
\rowcolor{HeaderSoft}
\textcolor{HeaderText}{\textbf{Benchmark}}
& \textcolor{HeaderText}{\textbf{Baseline}}
& \textcolor{HeaderText}{\textbf{\ours{}}}
& \textcolor{HeaderText}{\textbf{$\Delta$}}
& \textcolor{HeaderText}{\textbf{Layers}}
& \textcolor{HeaderText}{\textbf{Best route}} \\
\midrule

\modelbarroute{Qwen3-1.7B}
GSM8K & 43.44 & 69.52 & +26.08 & 28$\to$33 & \texttt{L0-L7 L3-L9 RJ} \\
MATH500 & 38.00 & 38.00 & +0.00 & 28 & \emph{standard forward pass} \\
MedQA & 40.53 & 41.32 & +0.79 & 28$\to$29 & \texttt{L0-L7 L7 RJ} \\
OpenBookQA & 68.60 & 75.00 & +6.40 & 28$\to$37 & \texttt{L0-L7 L8(+h-5) L9-L14 L9-L15 L13-L20 RJ} \\
SciQ & 92.60 & 93.30 & +0.70 & 28$\to$34 & \texttt{L0-L15 L10 L11 RJ} \\
MMLU-Pro & 23.62 & 28.25 & +4.62 & 28$\to$31 & \texttt{L0-L7 L5-L27 RJ} \\
SVAMP & 72.67 & 86.33 & +13.67 & 28$\to$33 & \texttt{L0-L8 L4 L5(+h-6) L6-L11 RJ} \\
ASDiv & 90.25 & 96.86 & +6.60 & 28$\to$28 & \texttt{L0-L7 L8(+h-1) L9-L27 RJ} \\
MAWPS & 57.88 & 66.35 & +8.46 & 28$\to$33 & \texttt{L0-L4 L5(+h-4) L6 L7 L3(+h-4) L4-L11 RJ} \\
GSM8K-Hard & 18.81 & 34.80 & +15.99 & 28$\to$32 & \texttt{L0-L2 L4-L7 L3-L9 RJ} \\
GSM8K-Plus & 29.20 & 53.60 & +24.40 & 28$\to$33 & \texttt{L0-L5 L1-L9 RJ} \\
MS-Arith & 26.00 & 91.00 & +65.00 & 28$\to$30 & \texttt{L0-L2 L4 L3-L7 L7 RJ} \\
Obj-Count & 61.00 & 75.00 & +14.00 & 28$\to$34 & \texttt{L0-L7 L7-L25 L24 L25 L24-L26 L26 L27 RJ} \\
\midrule
\modelbarroute{Qwen3-4B}
GSM8K & 74.83 & 81.35 & +6.52 & 36$\to$38 & \texttt{L0 L0-L5 L5 RJ} \\
MATH500 & 38.00 & 38.00 & +0.00 & 36 & \emph{standard forward pass} \\
MedQA & 62.77 & 62.77 & +0.00 & 36 & \emph{standard forward pass} \\
OpenBookQA & 90.80 & 90.80 & +0.00 & 36 & \emph{standard forward pass} \\
SciQ & 95.90 & 95.90 & +0.00 & 36 & \emph{standard forward pass} \\
MMLU-Pro & 44.00 & 47.12 & +3.12 & 36$\to$38 & \texttt{L0 L1 L0 RJ} \\
SVAMP & 88.00 & 90.67 & +2.67 & 36$\to$37 & \texttt{L0-L5 L5-L35 RJ} \\
ASDiv & 99.37 & 99.37 & +0.00 & 36 & \emph{standard forward pass} \\
MAWPS & 72.12 & 77.88 & +5.77 & 36$\to$35 & \texttt{L0-L7 L9(+h-3) L11(+h-6) L12-L20 L20-L35 RJ} \\
GSM8K-Hard & 31.66 & 40.13 & +8.46 & 36$\to$37 & \texttt{L0-L5 L5-L35 RJ} \\
GSM8K-Plus & 58.50 & 64.10 & +5.60 & 36$\to$37 & \texttt{L0-L5 L5-L35 RJ} \\
MS-Arith & 100.00 & 100.00 & +0.00 & 36 & \emph{standard forward pass} \\
Obj-Count & 93.00 & 93.00 & +0.00 & 36 & \emph{standard forward pass} \\
\midrule
\modelbarroute{Qwen3-8B}
GSM8K & 75.51 & 80.74 & +5.23 & 36$\to$38 & \texttt{L0 L0-L5 L5 RJ} \\
MATH500 & 42.00 & 42.00 & +0.00 & 36 & \emph{standard forward pass} \\
MedQA & 68.50 & 68.50 & +0.00 & 36 & \emph{standard forward pass} \\
OpenBookQA & 94.20 & 95.20 & +1.00 & 36$\to$38 & \texttt{L0-L2 L1 RJ} \\
SciQ & 96.00 & 96.50 & +0.50 & 36$\to$35 & \texttt{L0-L22 L24-L35 RJ} \\
MMLU-Pro & 51.00 & 51.00 & +0.00 & 36 & \emph{standard forward pass} \\
SVAMP & 89.33 & 89.33 & +0.00 & 36 & \emph{standard forward pass} (\texttt{L0-L35 RJ}) \\
ASDiv & 100.00 & 100.00 & +0.00 & 36 & \emph{standard forward pass} \\
MAWPS & 74.04 & 77.12 & +3.08 & 36$\to$37 & \texttt{L0 L1 L0-L10 L12-L14 RJ} \\
GSM8K-Hard & 37.62 & 40.75 & +3.13 & 36$\to$37 & \texttt{L0-L2 L2-L24 RJ} \\
GSM8K-Plus & 59.10 & 61.80 & +2.70 & 36$\to$38 & \texttt{L0 L1 L0 RJ} \\
MS-Arith & 98.89 & 99.44 & +0.56 & 36$\to$36 & \texttt{L0-L7 L9 L9-L35 RJ} \\
Obj-Count & 98.00 & 98.00 & +0.00 & 36 & \emph{standard forward pass} \\
\bottomrule
\end{tabular}
}
\caption{Best route found by \ours{} for each model and benchmark. }
\label{tab:best_routes_a}
\end{table*}

\begin{table*}[tp]
\centering
\small
\setlength{\tabcolsep}{5.0pt}
\renewcommand{\arraystretch}{1.10}
\arrayrulecolor{SoftRule}
\resizebox{\textwidth}{!}{
\begin{tabular}{p{\bmw}p{\numw}p{\numw}p{\delw}p{\layw}p{\routew}}
\toprule
\rowcolor{HeaderSoft}
\textcolor{HeaderText}{\textbf{Benchmark}}
& \textcolor{HeaderText}{\textbf{Baseline}}
& \textcolor{HeaderText}{\textbf{\ours{}}}
& \textcolor{HeaderText}{\textbf{$\Delta$}}
& \textcolor{HeaderText}{\textbf{Layers}}
& \textcolor{HeaderText}{\textbf{Best route}} \\
\midrule

\modelbarroute{Qwen3-14B (4-bit)}
GSM8K & 83.78 & 87.11 & +3.34 & 40$\to$46 & \texttt{L0-L6 L1(+h-5) L2 L3 RJ} \\
MATH500 & 41.00 & 47.00 & +6.00 & 40$\to$44 & \texttt{L0-L2 L4 L5 L1-L9 RJ} \\
MedQA & 73.53 & 73.53 & +0.00 & 40 & \emph{standard forward pass} (\texttt{L0-L39 RJ}) \\
OpenBookQA & 95.80 & 96.60 & +0.80 & 40$\to$41 & \texttt{L0-L2 L2-L24 RJ} \\
SciQ & 97.20 & 97.20 & +0.00 & 40 & \emph{standard forward pass} \\
MMLU-Pro & 56.62 & 56.62 & +0.00 & 40 & \emph{standard forward pass} \\
SVAMP & 92.00 & 92.33 & +0.33 & 40$\to$40 & \texttt{L0-L2 L2 L3(+h-2) L4-L7 L9-L39 RJ} \\
ASDiv & 99.06 & 99.69 & +0.63 & 40$\to$42 & \texttt{L0 L1 L0 L1 RJ} \\
MAWPS & 74.62 & 76.54 & +1.92 & 40$\to$45 & \texttt{L0-L5 L1-L4 RJ} \\
GSM8K-Hard & 41.38 & 43.26 & +1.88 & 40$\to$42 & \texttt{L0-L14 L13(+h-6) L14-L39 RJ} \\
GSM8K-Plus & 63.60 & 67.10 & +3.50 & 40$\to$44 & \texttt{L0-L4 L1 L2(+h-5) L3 RJ} \\
MS-Arith & 100.00 & 100.00 & +0.00 & 40 & \emph{standard forward pass} \\
Obj-Count & 100.00 & 100.00 & +0.00 & 40 & \emph{standard forward pass} \\
\midrule
\modelbarroute{Mixtral-8x7B (4-bit)}
GSM8K & 67.85 & 68.54 & +0.68 & 32$\to$33 & \texttt{L0-L7 L7-L31 RJ} \\
MATH500 & 23.00 & 27.00 & +4.00 & 32$\to$37 & \texttt{L0-L9 L5-L9 RJ} \\
MedQA & 55.62 & 57.19 & +1.57 & 32$\to$38 & \texttt{L0-L15 L10-L13 RJ} \\
OpenBookQA & 76.00 & 80.20 & +4.20 & 32$\to$43 & \texttt{L0-L15 L10-L13 L15 L16 L13-L22 L21-L31 RJ} \\
SciQ & 90.00 & 90.90 & +0.90 & 32$\to$38 & \texttt{L0-L15 L10 L11 RJ} \\
MMLU-Pro & 24.12 & 25.62 & +1.50 & 32$\to$33 & \texttt{L0-L6 L6(+h-5) L7-L31 RJ} \\
SVAMP & 78.67 & 78.67 & +0.00 & 32 & \emph{standard forward pass} \\
ASDiv & 91.82 & 95.91 & +4.09 & 32$\to$34 & \texttt{L0-L30 L29(+h-1) L30 L31 RJ} \\
MAWPS & 65.19 & 67.31 & +2.12 & 32$\to$36 & \texttt{L0-L7 L3(+h-4) L5(+h-6) L6-L11 RJ} \\
GSM8K-Hard & 33.54 & 33.54 & +0.00 & 32 & \emph{standard forward pass} (\texttt{L0-L31 RJ}) \\
GSM8K-Plus & 50.10 & 50.10 & +0.00 & 32 & \emph{standard forward pass} (\texttt{L0-L31 RJ}) \\
MS-Arith & 95.00 & 95.56 & +0.56 & 32$\to$36 & \texttt{L0-L4 L6 L7 L8(+h-1) L9 L5-L9 RJ} \\
Obj-Count & 65.00 & 65.00 & +0.00 & 32 & \emph{standard forward pass} (\texttt{L0-L31 RJ}) \\
\midrule
\modelbarroute{DeepSeek-R1-Distill-Llama-8B}
GSM8K & 20.17 & 36.54 & +16.38 & 32$\to$41 & \texttt{L0-L15 L10-L13 L15 L16 L13-L31 RJ} \\
MATH500 & 32.00 & 37.00 & +5.00 & 32$\to$32 & \texttt{L0-L3 L4(+h-1) L5-L23 RJ} \\
MedQA & 48.94 & 48.94 & +0.00 & 32 & \emph{standard forward pass} \\
OpenBookQA & 80.20 & 81.60 & +1.40 & 32$\to$31 & \texttt{L0-L18 L20-L31 RJ} \\
SciQ & 91.00 & 91.00 & +0.00 & 32 & \emph{standard forward pass} \\
MMLU-Pro & 33.62 & 33.62 & +0.00 & 32 & \emph{standard forward pass} \\
SVAMP & 36.33 & 60.00 & +23.67 & 32$\to$39 & \texttt{L0-L5 L5 L6 L1(+h-5) L2 L3 RJ} \\
ASDiv & 46.23 & 86.79 & +40.57 & 32$\to$43 & \texttt{L0-L5 L5 L6 L1(+h-5) L2-L16 L13-L31 RJ} \\
MAWPS & 35.38 & 58.46 & +23.08 & 32$\to$36 & \texttt{L0-L5 L5 L6 L4(+h-1) L5-L11 RJ} \\
GSM8K-Hard & 20.69 & 21.63 & +0.94 & 32$\to$34 & \texttt{L0-L7 L6(+h-5) L7-L18 L19(+h-3) L20-L31 RJ} \\
GSM8K-Plus & 46.00 & 46.00 & +0.00 & 32 & \emph{standard forward pass} (\texttt{L0-L31 RJ}) \\
MS-Arith & 36.67 & 86.67 & +50.00 & 32$\to$42 & \texttt{L0-L15 L11-L13 L9-L20 RJ} \\
Obj-Count & 34.00 & 70.00 & +36.00 & 32$\to$33 & \texttt{L0-L18 L18-L22 L24(+h-6) L25-L27 L27-L31 RJ} \\
\bottomrule
\end{tabular}
}
\caption{Best route found by \ours{} for each model and benchmark}
\label{tab:best_routes_b}
\end{table*}

\begin{table*}[tp]
\centering
\small
\setlength{\tabcolsep}{5.0pt}
\renewcommand{\arraystretch}{1.10}
\arrayrulecolor{SoftRule}
\resizebox{\textwidth}{!}{
\begin{tabular}{p{\bmw}p{\numw}p{\numw}p{\delw}p{\layw}p{\routew}}
\toprule
\rowcolor{HeaderSoft}
\textcolor{HeaderText}{\textbf{Benchmark}}
& \textcolor{HeaderText}{\textbf{Baseline}}
& \textcolor{HeaderText}{\textbf{\ours{}}}
& \textcolor{HeaderText}{\textbf{$\Delta$}}
& \textcolor{HeaderText}{\textbf{Layers}}
& \textcolor{HeaderText}{\textbf{Best route}} \\
\midrule

\modelbarroute{Llama-3.2-3B-Instruct}
GSM8K & 75.36 & 76.04 & +0.68 & 28$\to$28 & \texttt{L0 L1 L2(-0.25h-1) RJ} \\
MATH500 & 44.00 & 44.00 & +0.00 & 28 & \emph{standard forward pass} (\texttt{L0-L2 RJ}) \\
MedQA & 57.42 & 57.42 & +0.00 & 28 & \emph{standard forward pass} (\texttt{L0 RJ}) \\
OpenBookQA & 77.00 & 77.00 & +0.00 & 28 & \emph{standard forward pass} (\texttt{L0 RJ}) \\
SciQ & 89.40 & 89.40 & +0.00 & 28 & \emph{standard forward pass} \\
MMLU-Pro & 33.25 & 33.25 & +0.00 & 28 & \emph{standard forward pass} (\texttt{L0 RJ}) \\
SVAMP & 84.33 & 84.33 & +0.00 & 28 & \emph{standard forward pass} \\
ASDiv & 96.23 & 96.23 & +0.00 & 28 & \emph{standard forward pass} \\
MAWPS & 64.23 & 64.23 & +0.00 & 28 & \emph{standard forward pass} \\
GSM8K-Hard & 26.65 & 26.65 & +0.00 & 28 & \emph{standard forward pass} \\
GSM8K-Plus & 58.30 & 58.30 & +0.00 & 28 & \emph{standard forward pass} \\
MS-Arith & 97.78 & 97.78 & +0.00 & 28 & \emph{standard forward pass} \\
Obj-Count & 57.00 & 63.00 & +6.00 & 28$\to$31 & \texttt{L0-L7 L5-L27 RJ} \\
\bottomrule
\end{tabular}
}
\caption{Best route found by \ours{} for each model and benchmark.}
\label{tab:best_routes_c}
\end{table*}

\definecolor{posgreen}{HTML}{15803D} 
\definecolor{negred}{HTML}{B91C1C}   
\newcommand{\good}[1]{\textcolor{posgreen}{#1}}
\newcommand{\bad}[1]{\textcolor{negred}{#1}}
\definecolor{SubtotalBand}{HTML}{F9FAFB}

\newcommand{\subhead}[1]{%
  \multicolumn{7}{@{}l}{\textit{#1}}\\[1pt]%
}
\begin{table*}[tp]
\centering
\begin{tabular}{l ccc ccc}
\toprule
\rowcolor{HeaderSoft}
\textcolor{HeaderText}{} &
\multicolumn{3}{c}{\textcolor{HeaderText}{\textbf{Qwen3-1.7B}}} &
\multicolumn{3}{c}{\textcolor{HeaderText}{\textbf{Qwen3-4B}}} \\
\rowcolor{HeaderSoft}
\textcolor{HeaderText}{\textbf{Benchmark}} &
\textcolor{HeaderText}{Base.} & \textcolor{HeaderText}{\ours{}} & \textcolor{HeaderText}{$\Delta$} &
\textcolor{HeaderText}{Base.} & \textcolor{HeaderText}{\ours{}} & \textcolor{HeaderText}{$\Delta$} \\
\midrule

\rowcolor{HeaderSoft}
\multicolumn{7}{l}{\textbf{In-Domain}} \\
MAWPS      & 58.46 & 62.95\std{3.95} & \good{+4.49}  & 71.73 & 72.24\std{0.48} & \good{+0.51} \\
ASDiv      & 91.51 & 94.03\std{1.78} & \good{+2.52}  & 99.06 & 99.16\std{0.15} & \good{+0.10} \\
SVAMP      & 74.00 & 81.44\std{5.27} & \good{+7.44}  & 88.33 & 89.33\std{0.98} & \good{+1.00} \\
MultiArith & 95.00 & 96.85\std{1.31} & \good{+1.85}  & 99.44 & 99.81\std{0.26} & \good{+0.37} \\
Obj-Count  & 54.00 & 58.67\std{3.30} & \good{+4.67}  & 90.00 & 90.67\std{1.70} & \good{+0.67} \\
MATH500    & 38.00 & 36.33\std{3.09} & \bad{-1.67}   & 34.00 & 36.00\std{3.56} & \good{+2.00} \\
OpenBookQA & 67.60 & 72.73\std{4.72} & \good{+5.13}  & 91.80 & 90.20\std{1.18} & \bad{-1.60}  \\
SciQ       & 93.40 & 91.30\std{2.24} & \bad{-2.10}   & 95.80 & 95.13\std{0.53} & \bad{-0.67}  \\
\addlinespace[1pt]
\rowcolor{SubtotalBand}
\textbf{Overall} & \textbf{71.50} & \textbf{74.29}\std{\textbf{2.07}} & \good{\textbf{+2.79}} &
                   \textbf{83.77} & \textbf{84.07}\std{\textbf{0.74}} & \good{\textbf{+0.30}} \\
\midrule

\rowcolor{HeaderSoft}
\multicolumn{7}{l}{\textbf{Out-of-Domain}} \\
GSM8K      & 43.75 & 60.85\std{12.18} & \good{+17.11} & 81.80 & 80.92\std{1.64} & \bad{-0.88}  \\
GSM-Hard   & 17.24 & 26.23\std{6.85}  & \good{+8.99}  & 33.23 & 39.39\std{4.51} & \good{+6.17} \\
GSM-Plus   & 31.10 & 45.10\std{10.01} & \good{+14.00} & 57.00 & 61.70\std{3.32} & \good{+4.70} \\
MMLU-Pro   & 23.75 & 27.42\std{2.66}  & \good{+3.67}  & 45.50 & 45.38\std{0.27} & \bad{-0.13}  \\
MedQA      & 39.04 & 40.38\std{3.19}  & \good{+1.34}  & 62.92 & 58.65\std{3.05} & \bad{-4.27}  \\
\addlinespace[1pt]
\rowcolor{SubtotalBand}
\textbf{Overall} & \textbf{30.98} & \textbf{40.00}\std{\textbf{6.39}} & \good{\textbf{+9.02}} &
                   \textbf{56.09} & \textbf{57.21}\std{\textbf{0.82}} & \good{\textbf{+1.12}} \\
\bottomrule
\end{tabular}

\begin{minipage}{\linewidth}
\footnotesize
\vspace{4pt}
\hspace*{12em}%
\begin{tabular}{ l l l @{}}
\textbf{Model} & \textbf{Layers} & \textbf{Best route }\\
\textbf{Qwen3-1.7B:} & 28$\to$34 & \texttt{L0-L4 L5(+h-4) L6-L15 L10-L11 RJ} \\
\textbf{Qwen3-4B:}   & 36$\to$37 & \texttt{L0-L5 L5 L6 RJ}
\end{tabular}
\end{minipage}
\caption{Performance of \ours{} evaluated across all in-domain and out-of-domain
benchmarks for each model. Reported accuracies are averaged over 3 seeds; the layer range and route
shown below the table are from one representative seed, given as an illustrative
example.}
\label{tab:one_model_one_route_benchmarks}
\end{table*}

\paragraph{Selected Route Composition}
Table~\ref{tab:best_path_action_presence} summarizes which nonstandard actions
appear in the final top-1 routes selected by \ours{} and which Viterbi rank is
ultimately chosen after validation. Move actions are common across model
families, while Add appears more selectively; every valid route eventually uses
\textsc{Rejoin}, so it is omitted from the table.

\begin{center}
\centering
\small
\setlength{\tabcolsep}{4.5pt}
\renewcommand{\arraystretch}{1.12}
\arrayrulecolor{SoftRule}
\resizebox{\columnwidth}{!}{
\begin{tabular}{lrrrrr}
\toprule
\rowcolor{HeaderSoft}
\textcolor{HeaderText}{\textbf{Model}}
& \textcolor{HeaderText}{\textbf{Add (\%)}}
& \textcolor{HeaderText}{\textbf{Move (\%)}}
& \textcolor{HeaderText}{\textbf{Both (\%)}}
& \textcolor{HeaderText}{\textbf{Len. $\Delta$ (\%)}}
& \textcolor{HeaderText}{\textbf{Mean Rank}} \\
\midrule
Qwen3-1.7B            & 54.3 & 65.7 & 25.7 & $+9.5$  & 1.81 \\
Qwen3-4B              & 15.4 & 65.4 &  7.7 & $+1.9$  & 2.23 \\
Qwen3-8B              &  7.7 & 65.4 &  3.8 & $+2.3$  & 2.64 \\
Qwen3-14B (4-bit)     & 26.9 & 69.2 & 19.2 & $+4.4$  & 2.46 \\
Llama-3.2-3B          &  0.0 & 38.9 &  0.0 & $+3.6$  & 1.72 \\
Mixtral-8x7B (4-bit)  & 45.5 & 68.2 & 31.8 & $+7.9$  & 2.62 \\
DeepSeek-R1-8B        & 31.2 & 50.0 & 21.9 & $+10.7$ & 2.05 \\
\bottomrule
\end{tabular}}
\captionof{table}{Frequency with which selected top-1 routes contain each action at
least once, together with relative route-length change and the average raw
Viterbi rank selected after validation among the final top-5 candidates (rank 1 being the most probable).
``Both'' denotes routes that contain at least one Add and at least one Move.
Length change is
$100(\overline{L}_{\mathrm{ours}}-\overline{L}_{\mathrm{base}})/
\overline{L}_{\mathrm{base}}$, where $L$ is the number of executed layer applications.
}
\label{tab:best_path_action_presence}
\end{center}

\FloatBarrier

\section{Experimental Details}
\label{app:experimental_details}

\paragraph{Connection to CEM.} Our search procedure can be viewed as a structured variant of the Cross-Entropy Method. The sampling distribution is a masked Markov policy over routing programs. Each iteration samples candidate programs, evaluates them on training data, forms an elite or soft-elite distribution using accuracy-based weights, and updates the Markov transition table by weighted maximum likelihood. Thus, the update minimizes cross-entropy between the learned policy family and the empirical distribution induced by high-performing sampled routes, while feasibility masks and smoothing adapt CEM to the constrained routing-program space.

\paragraph{Splits and Decoding}
For each model, benchmark, and random seed, we learn a separate task-level route
policy while keeping all pretrained model weights frozen. Each run requests
1,000 training examples and a held-out validation split for route selection
(100 examples in the standard runs and 200 examples in the baseline-anchored
4-bit/non-Qwen runs; smaller benchmark train pools are split proportionally by
the loader). The final test score is computed after route selection using the
available test split or held-out remainder, with no test subsampling unless
explicitly configured. We report means and standard deviations over completed
seeds from \(\{1,2,42\}\). Generation is deterministic greedy decoding with
temperature \(0\), top-\(p=1.0\), top-\(k=0\), at most 400 thinking tokens, and
at most 128 answer tokens.

\paragraph{\ours{} Hyperparameters}
The main experiments use 10 search iterations with 30 candidate programs per
iteration. Training examples are partitioned into iteration chunks; each
iteration also replays one tenth \((\lambda_{\mathrm{rep}}=0.1)\) of each previous chunk. The Markov state uses
three budget-phase bins, incoming layer-delta context, and operator context,
but not the previous-layer context. The local move radius is \(r=5\). The
program action space contains identity layer moves, add-and-apply actions, and
\textsc{Rejoin}; subtraction and bare merge actions are disabled in the main
\ours{} runs. Add actions draw from a history window of six previous hidden
states and use coefficient \(\gamma=1.0\). The initial policy sets
\(\varepsilon_{\mathrm{loc}}=0.2\) for local layer moves and \textsc{Rejoin},
and \(\varepsilon_{\mathrm{op}}=0.0867\) for add-and-apply actions. The transition update uses smoothing \(\alpha=0.0397\),
softmax sharpness \(\beta=1.8262\), and an elite set of the top five candidates
per iteration. Route sampling temperature is 1.0. Search stops early when all
sampled programs become semantically identical after expanding
\textsc{Rejoin}; otherwise it stops at the 10-iteration cap.

\begin{table}[t]
\centering
\small
\setlength{\tabcolsep}{4.5pt}
\renewcommand{\arraystretch}{1.12}
\arrayrulecolor{SoftRule}
\resizebox{\columnwidth}{!}{
\begin{tabular}{lcc}
\toprule
\rowcolor{HeaderSoft}
\textcolor{HeaderText}{\textbf{Model group}}
& \textcolor{HeaderText}{\textbf{Max route len.}}
& \textcolor{HeaderText}{\textbf{Final set}} \\
\midrule
Qwen3-1.7B, Qwen3-4B, Qwen3-8B, Llama-3.2-3B & 40 & top-5 Viterbi \\
Qwen3-14B (4-bit)                            & 52 & baseline + 4 Viterbi neighbors \\
Mixtral-8x7B (4-bit), DeepSeek-R1-8B         & 44 & baseline + 4 Viterbi neighbors \\
\bottomrule
\end{tabular}}
\caption{Model-specific route-length caps and final candidate construction.
Viterbi decoding returns the highest-probability raw programs under the masked
Markov policy. For final evaluation we deduplicate programs that execute the
same route after expanding \textsc{Rejoin}; in baseline-anchored runs, the
standard route is inserted as the selection floor and the remaining candidates
are the highest-probability nonstandard neighbors.}
\label{tab:experimental_hparams}
\end{table}

\paragraph{Final Selection and Baselines}
After the final Markov update, we decode five final candidates with top-\(k\)
Viterbi, evaluate them on the validation split, and select by validation
accuracy, breaking ties by training accuracy and then shorter route length. The
reported test score is the selected route's held-out test accuracy. The Dr.LLM baseline trains one lightweight router per
benchmark from MCTS-generated route labels. For Dr.LLM+Ext., we keep the
Dr.LLM training procedure but replace its action space with the extended
\ours{} program action space. The Program-Dr.LLM runs use 100 MCTS-labeled
training examples, 10 MCTS simulations per example, exploration constant 1.8,
length penalty 3.0, router hidden dimension 128, eight windows, 15 epochs,
AdamW learning rate \(10^{-3}\), weight decay 0.01, gradient accumulation 16,
and focal-loss \(\gamma=2.0\).

\paragraph{Exactness of top-$k$ decoding}
The five final candidates are the \emph{exact} $k$ highest-probability programs
under the learned policy, not a beam or sampling approximation. Exactness is not
automatic here: because the state conditions on the incoming layer-delta (the
signed gap to the previously applied layer) and on the operator context, the
policy is \emph{not} Markov in the current layer alone, so a decoder that merged
all paths reaching a given layer and budget phase would be approximate. We
instead run the top-$k$ dynamic program over the augmented state: current layer,
budget phase, incoming delta, and operator context, keying on the previously
applied layer (which fixes the incoming delta exactly) alongside the current
layer, length, and operator context, so partial paths with differing
continuations are never merged. The augmentation is lossless and cheap: the
incoming delta is a one-step feature bounded by the local radius $r$, so it
enlarges the state space by at most a factor $2r+1$ and needs no longer-range
history, and since program length grows by one at every step the reachable set
is a finite DAG on which retaining the $k$ best continuations per state yields
the exact global top-$k$. After decoding we deduplicate programs that execute the
same route once \textsc{Rejoin} is expanded, keeping the most probable
representative of each route.

\FloatBarrier

\section{Inference-Time Cost}
\label{app:inference_cost}

\begin{figure*}[t]
\centering
\includegraphics[width=\textwidth]{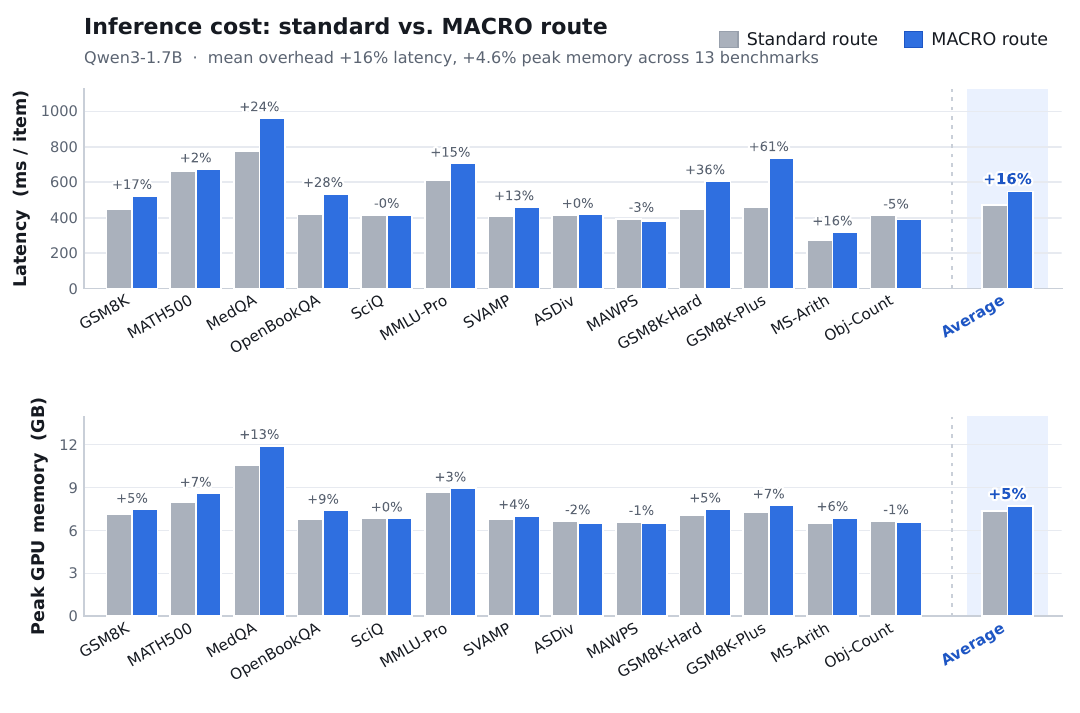}
\caption{Per-item decoding latency (top) and peak GPU memory (bottom) of the
standard forward pass (gray) versus the \ours{} route (blue) on
Qwen3-1.7B, under the same two-stage 400/128-token generation budget with batched
greedy decoding on a single A40 over 64 test items per benchmark. The percentage
above each pair is the route's
overhead relative to the standard route; the rightmost group averages over all
thirteen benchmarks (\(+16.1\%\) latency, \(+4.6\%\)/\(335\) MB memory). Peak
memory is the maximum allocated during generation, not compute utilization.}
\label{fig:inference_cost}
\end{figure*}

Because \ours{} routes revisit, repeat, or occasionally drop layers, the executed
route changes the per-token cost of decoding relative to the standard forward
pass. Figure~\ref{fig:inference_cost} profiles this on Qwen3-1.7B, comparing the
standard route with the \ours{} route on decoding latency and peak
GPU memory. The overhead is modest and uneven: it is largest where the selected
route is longest, negligible where the route stays close to baseline length, and
occasionally negative, faster and lighter, when the route is shorter than the
standard pass. Since these are the routes that deliver the accuracy gains of
Table~\ref{tab:best_routes_a}, \ours{} trades a small, route-dependent compute
premium for higher accuracy at fixed model weights.

\section*{Use of LLM Assistance}
Large language models were used only as auxiliary aids for editing, grammar,
presentation, and implementation-code support. The authors made all substantive
technical contributions, experimental design decisions, analyses,
interpretations, and final research judgments.

\FloatBarrier

\end{document}